\documentclass{bmvc2k}
\usepackage{graphicx}
\usepackage{amsmath}
\usepackage{amssymb}
\usepackage{booktabs}
\usepackage{float}
\usepackage{multirow}
\usepackage{multicol}
\usepackage{subfigure}
\usepackage{algorithm}
\usepackage{algorithmic}
\usepackage{breakcites}
\usepackage{xcolor}
\usepackage{color}

\title{Copy-Pasting Coherent Depth Regions Improves Contrastive Learning for Urban-Scene Segmentation}
\addauthor{Liang Zeng}{liangzeng1996@outlook.com}{1}
\addauthor{Attila Lengyel}{A.Lengyel@tudelft.nl}{1}
\addauthor{Nergis Tömen}{N.Tomen@tudelft.nl}{1}
\addauthor{Jan van Gemert}{J.C.vanGemert@tudelft.nl}{1}
\addinstitution{
 Delft University of Technology\\
 Delft, The Netherlands
}

\runninghead{Zeng et al}{Copy-Paste and Depth Improve Contrastive Learning}

\begin{document}

\maketitle

\begin{abstract}
In this work, we leverage estimated depth to boost self-supervised contrastive learning for segmentation of urban scenes, where unlabeled videos are readily available for training self-supervised depth estimation. We argue that the semantics of a coherent group of pixels in 3D space is self-contained and invariant to the contexts in which they appear. We group coherent, semantically related pixels into coherent depth regions given their estimated depth and use copy-paste to synthetically vary their contexts. In this way, cross-context correspondences are built in contrastive learning and a context-invariant representation is learned. For unsupervised semantic segmentation of urban scenes, our method surpasses the previous state-of-the-art baseline by $+7.14\%$ in mIoU on Cityscapes and $+6.65\%$ on KITTI. For fine-tuning on Cityscapes and KITTI segmentation, our method is competitive with existing models, yet, we do not need to pre-train on ImageNet or COCO, and we are also more computationally efficient. Our code is available on {\footnotesize \url{https://github.com/LeungTsang/CPCDR}}.

\end{abstract}

\section{Introduction}
The lack of labeled data is a bottleneck for deep-learning-based computer vision techniques due to the cost of annotating~\cite{ILSVRC15,DBLP:conf/cvpr/CordtsORREBFRS16}. To this end, researchers have established the "self-supervised pre-training then fine-tuning" paradigm which can use unlabeled data. One of the most successful approaches is self-supervised contrastive learning~\cite{DBLP:conf/cvpr/HadsellCL06,DBLP:conf/icml/ChenK0H20,DBLP:conf/nips/ChenKSNH20,DBLP:conf/cvpr/He0WXG20,DBLP:journals/corr/abs-2003-04297}.

\begin{figure}[H]
\centering
\begin{tabular}{cc}
\centering
\bmvaHangBox{\includegraphics[width=2.5cm]{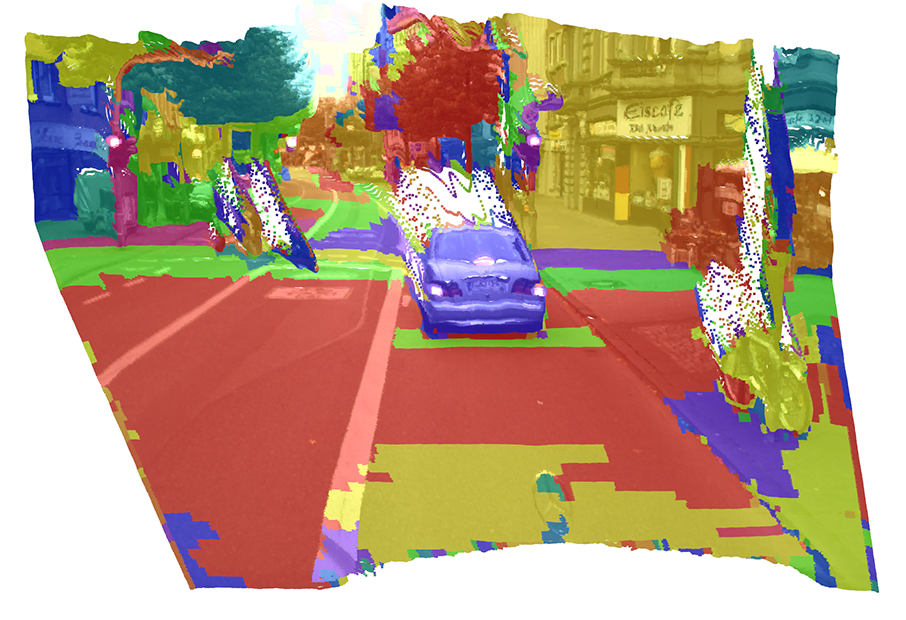}}&
\bmvaHangBox{\includegraphics[width=7.5cm]{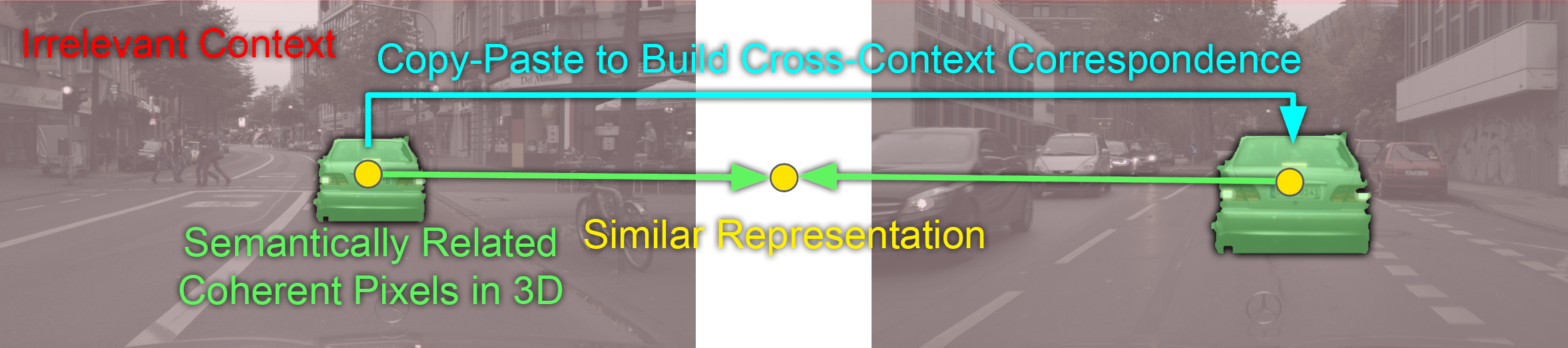}}\\
(a) & (b)
\end{tabular}
\caption{(a) We group coherent, semantically related pixels into coherent depth regions as a prior for contrastive learning given estimated depth. We argue that the semantics of the coherent pixels are self-contained regardless of different contexts. (b) We copy-paste the regions to build cross-context correspondences. The corresponding representations are pulled together by contrastive learning.} 

\label{intro}
\end{figure}

We explore how to use depth to aid self-supervised contrastive learning for urban-scene segmentation~\cite{DBLP:conf/cvpr/CordtsORREBFRS16,Geiger2013IJRR}. Applying contrastive learning to complex, non-object-centric urban scenes for segmentation is a non-trivial and often overlooked research topic. We argue that the semantic relatedness of pixels correlates with their coherence in 3D space. Therefore, we use estimated depth to group coherent, semantically related pixels into coherent depth regions. We then apply a copy-paste data augmentation strategy~\cite{DBLP:conf/cvpr/GhiasiCSQLCLZ21} to artificially vary the contexts in which the grouped coherent pixels appear and build cross-context correspondences. These correspondences are used as positive pairs in contrastive learning. As self-supervised depth estimation on urban scenes is well-addressed in literature~\cite{DBLP:conf/cvpr/ZhouBSL17,DBLP:conf/iccv/GodardAFB19,DBLP:journals/corr/abs-2003-06620} depth can be extracted from images freely.

\begin{figure}[H]
\centering
\begin{tabular}{ccc}
\centering
\bmvaHangBox{\includegraphics[width=3.5cm]{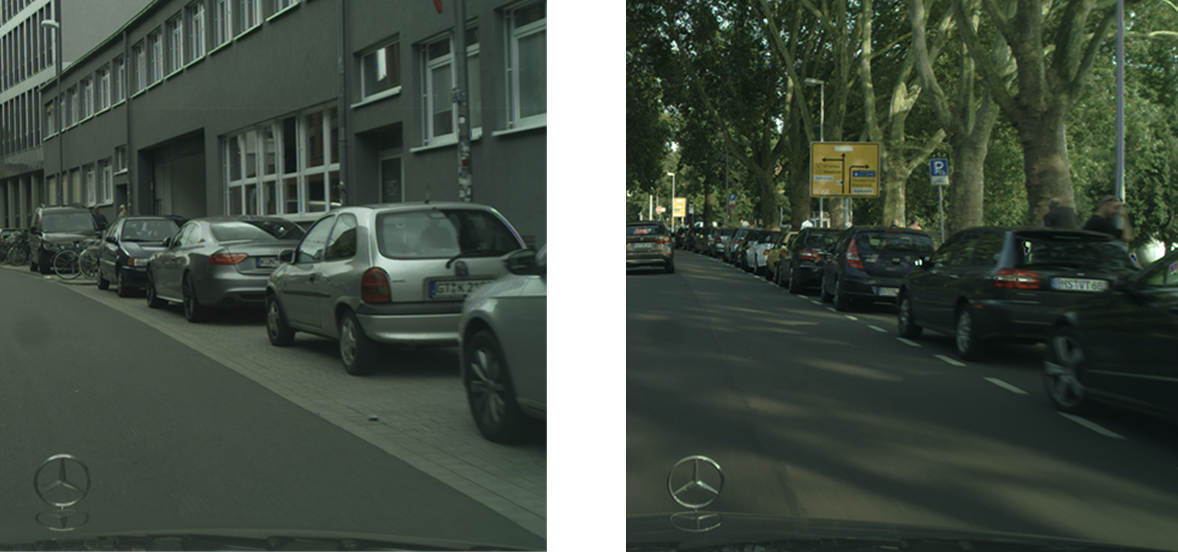}}&
\bmvaHangBox{\includegraphics[width=3.5cm]{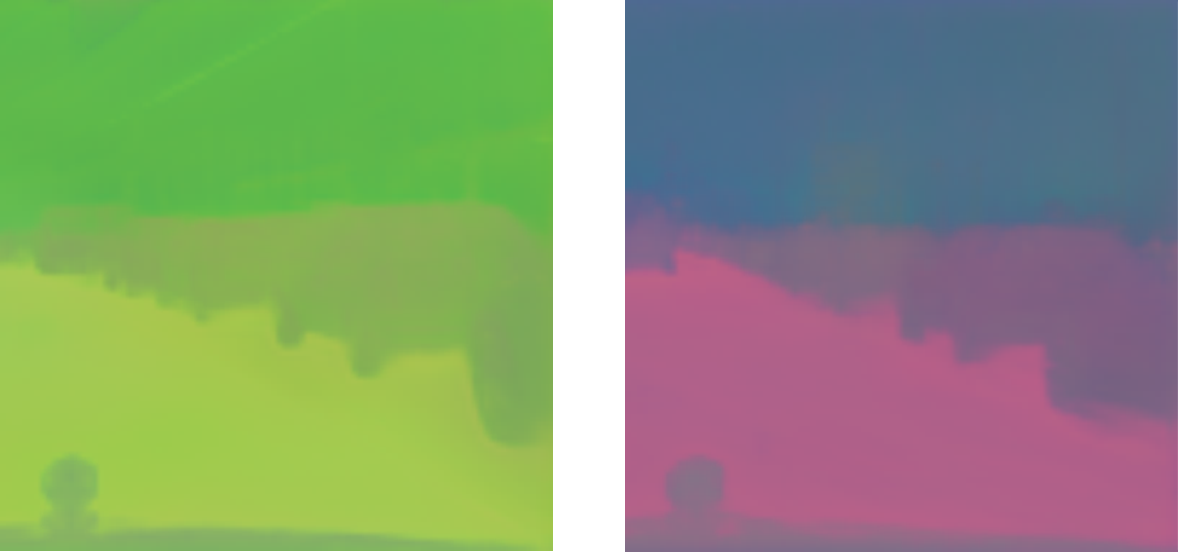}}&
\bmvaHangBox{\includegraphics[width=3.5cm]{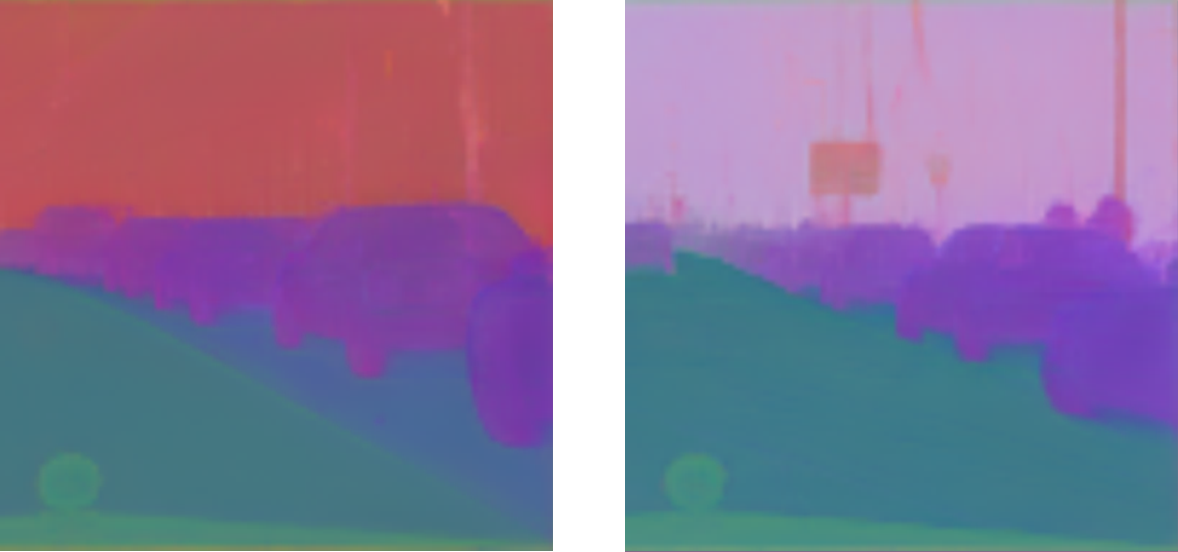}}\\
(a) Cars in different contexts&(b) w/o copy-paste & (c)with copy-paste
\end{tabular}

\caption{Feature maps are visualized as RGB images by PCA reduction~\cite{DBLP:conf/eccv/VondrickSFGM18}. Copy-paste is vital for learning object-specific representations invariant across different contexts.}

\label{cp}
\end{figure}

We explore if a group of coherent pixels is invariant over different contexts and if such invariance benefits representation learning. Deep models for visual recognition are sensitive to contexts~\cite{DBLP:journals/corr/abs-2205-02887,DBLP:conf/cvpr/ZhangTK20, DBLP:conf/cvpr/BarneaB19}, where spurious correlations with the contexts learn non-existing shortcuts~\cite{DBLP:journals/natmi/GeirhosJMZBBW20,DBLP:conf/eccv/NorooziF16}. This shortcut comes from the large receptive field of modern neural networks~\cite{He_2016_CVPR}  and varying contexts by copy-pasting coherent regions in various scenes encourages the learned representations to be invariant to contexts. In Figure~\ref{cp} we visualize the effect of our approach.

Our contributions are summarized as follows. (1) We demonstrate that copy-paste augmentation of coherent depth regions is important for unsupervised contrastive learning of object-specific and context-invariant representations. (2) Our method allows effective pre-
training on urban scene data, without the need for an additional large-scale dataset. (3) For unsupervised semantic segmentation, our method outperforms the previous state-of-the-art baseline by a significant $+7.14\%$ in mIoU on Cityscapes~\cite{DBLP:conf/cvpr/CordtsORREBFRS16} and $+6.65\%$ on KITTI~\cite{Geiger2013IJRR}. (4) For fine-tuning, our method allows effective pre-training on urban scenes with a single GPU. Our model achieves competitive segmentation performance on Cityscapes~\cite{DBLP:conf/cvpr/CordtsORREBFRS16} and KITTI~\cite{Geiger2013IJRR} to existing models that are pre-trained on ImageNet~\cite{ILSVRC15} or COCO~\cite{DBLP:conf/eccv/LinMBHPRDZ14} with more than 8 GPUs.
    
\section{Related Work}

\paragraph{Self-supervised Contrastive Learning}
Self-supervised pre-training has gained significant attention in computer vision thanks to its ability to use unlabeled data which is generally available in large quantities. Many self-supervised learning tasks have been explored~\cite{DBLP:conf/iccv/WangG15,DBLP:conf/iccv/DoerschGE15,DBLP:conf/eccv/NorooziF16,DBLP:conf/eccv/ZhangIE16,DBLP:conf/cvpr/PathakGDDH17,DBLP:conf/iclr/GidarisSK18,DBLP:journals/corr/abs-2111-06377}, of which contrastive learning~\cite{DBLP:conf/cvpr/HadsellCL06} currently most prevalent. 

The design of contrastive learning, especially the definition of positive and negative samples, depends on the data and task at hand. Pioneering works~\cite{DBLP:conf/icml/ChenK0H20,DBLP:conf/nips/ChenKSNH20,DBLP:conf/cvpr/He0WXG20,DBLP:journals/corr/abs-2003-04297,DBLP:conf/iccv/ChenXH21,DBLP:conf/cvpr/ChenH21,DBLP:conf/nips/CaronMMGBJ20,DBLP:conf/icml/ZbontarJMLD21,DBLP:conf/nips/Tian0PKSI20} are mostly based on instance discrimination on ImageNet~\cite{ILSVRC15}. Positive pairs are generated by applying different transformations to the same image, while negative pairs consist of different samples. For dense prediction~\cite{7298965}, VADeR~\cite{DBLP:conf/nips/PinheiroABGC20}, DenseCL~\cite{DBLP:conf/cvpr/WangZSKL21} and PixPro~\cite{DBLP:conf/cvpr/XieL00L021} learn dense visual representations by performing instance discrimination at pixel-level. For object detection on complex scenes~\cite{DBLP:conf/nips/RenHGS15}, object-level instance discrimination is possible if object locations are known~\cite{DBLP:conf/cvpr/SelvarajuD0N21,xie2021unsupervised}. For segmentation additional similarity can be defined among certain regions on images, such as classic hierarchical pixel grouping~\cite{DBLP:conf/nips/ZhangM20}, auxiliary labels~\cite{zhang2021looking} and salient object estimation~\cite{DBLP:conf/iccv/GansbekeVGG21}.

We adopt SwAV~\cite{DBLP:conf/nips/CaronMMGBJ20} to perform contrastive learning as it only needs positive samples. We extend it to learn dense visual representation as our goal is segmentation. The positive samples can be defined at pixel-level or region-level. We compute precise pixel correspondences with respect to the geometric transformations similarly as in PixPro~\cite{DBLP:conf/cvpr/XieL00L021} and VADeR~\cite{DBLP:conf/nips/PinheiroABGC20}, but our geometric transformations involve copy-paste in addition. Region-level positive samples are sampled from pixel grouping results as recent works~\cite{DBLP:conf/nips/ZhangM20,zhang2021looking,DBLP:conf/iccv/GansbekeVGG21,DBLP:conf/iccv/HenaffKAOVC21}, but our pixel grouping is based on the novel 3D coherence given the depth estimation.

\paragraph{Copy-Paste Data Augmentation}

Copy-Paste is a well-studied augmentation in supervised learning~\cite{DBLP:conf/iccv/YunHCOYC19,DBLP:conf/iccv/FangSWGLL19,DBLP:conf/cvpr/GhiasiCSQLCLZ21}. Copy-paste alike augmentations can also be used for self-supervised contrastive learning as long as we know what to copy-paste. DiLo~\cite{DBLP:conf/aaai/ZhaoWLL21} swaps salient objects from images while RegionCL~\cite{DBLP:journals/corr/abs-2111-12309} and CP2~\cite{DBLP:journals/corr/abs-2203-11709} swap random rectangular patches. However, the assumption that the extracted objects or patches are semantically equivalent to the original images is only valid on simple object-centric datasets~\cite{ILSVRC15}. 

Our work utilizes depth to define semantically related regions for copy-paste which improves the learning of dense visual representations in complex urban scenes.

\paragraph{Self-Supervised Depth Estimation}
Self-supervised depth estimation on videos has gained impressive success~\cite{DBLP:conf/cvpr/ZhouBSL17,DBLP:conf/iccv/GodardAFB19,DBLP:journals/corr/abs-2003-06620}. As video clips are generally available in urban-scene datasets~\cite{DBLP:conf/cvpr/CordtsORREBFRS16,Geiger2013IJRR}, depth estimation may help supervised segmentation in different ways~\cite{DBLP:conf/cvpr/HoyerDCKSG21}, including jointly training, data selection and DepthMix data augmentation.

In our work, estimated depth is used for grouping semantically related pixels and generating augmented input samples for self-supervised learning. DepthMix~\cite{DBLP:conf/cvpr/HoyerDCKSG21} is adopted to enhance copy-paste augmentation. 

\section{Method}
\begin{figure}[H]
  \centering
    \includegraphics[width=1\linewidth]{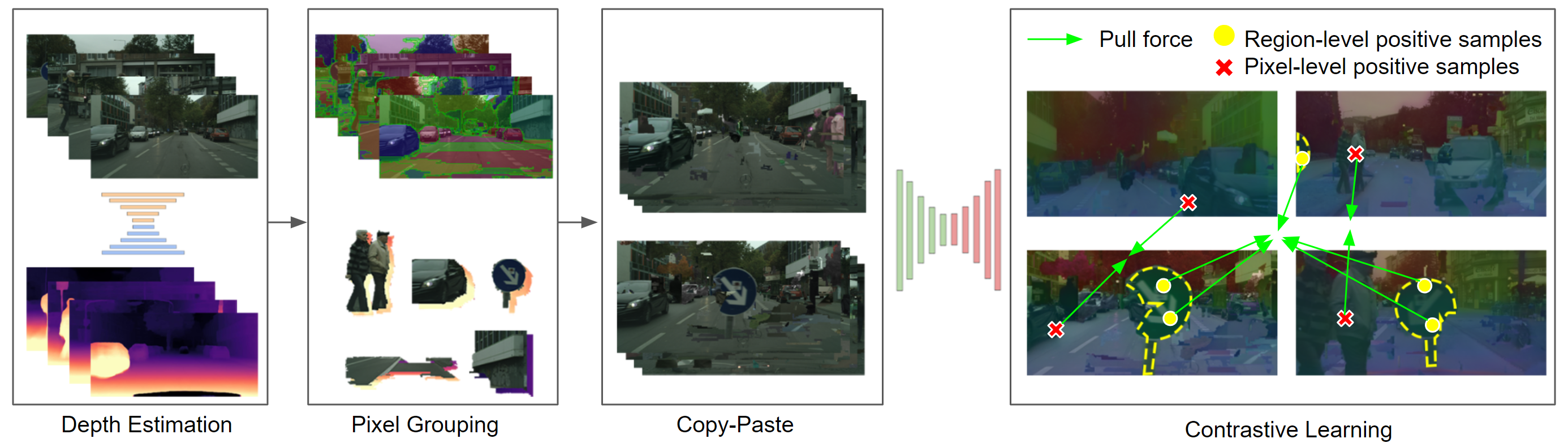}
    \caption{Our method consists of four steps. 1. Training a depth estimator on video clips by self-supervision. 2. Grouping pixels coherent in 3D space given the depth. 3. Building cross-context correspondences by copy-paste. 4. Pulling together the representations of corresponding pixels and regions using contrastive learning.}
    \label{pipeline}
\end{figure}

We illustrate the pipeline of our method which consists of four steps, as shown in Figure~\ref{pipeline}.

\subsection{Self-Supervised Depth Estimation}

To obtain depth, we take advantage of the recent success of self-supervised depth estimation on monocular videos~\cite{DBLP:conf/cvpr/ZhouBSL17,DBLP:conf/iccv/GodardAFB19,DBLP:journals/corr/abs-2003-06620}. In our framework, we adopt Monodepth2~\cite{DBLP:conf/iccv/GodardAFB19} in virtue of its simplicity and effectiveness.

\subsection{Coherent Pixels Grouping}

We propose a heuristic algorithm to group semantically related pixels that are coherent in 3D space into coherent depth regions based on the estimated depth and horizontal pattern of urban-scene images. 

\begin{figure}
\centering
\begin{tabular}{cc}
\centering
\bmvaHangBox{\includegraphics[width=5cm]{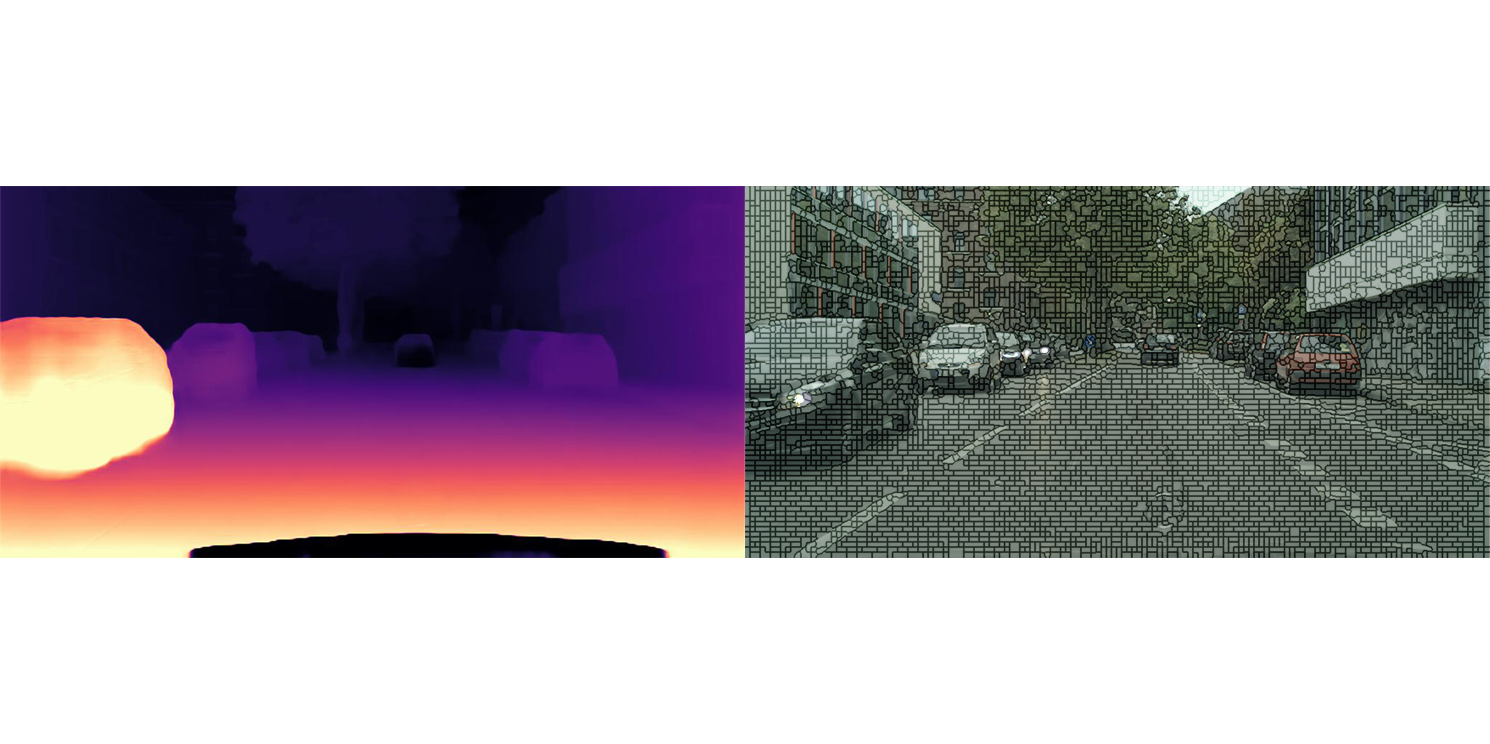}}&
\bmvaHangBox{\includegraphics[width=5cm]{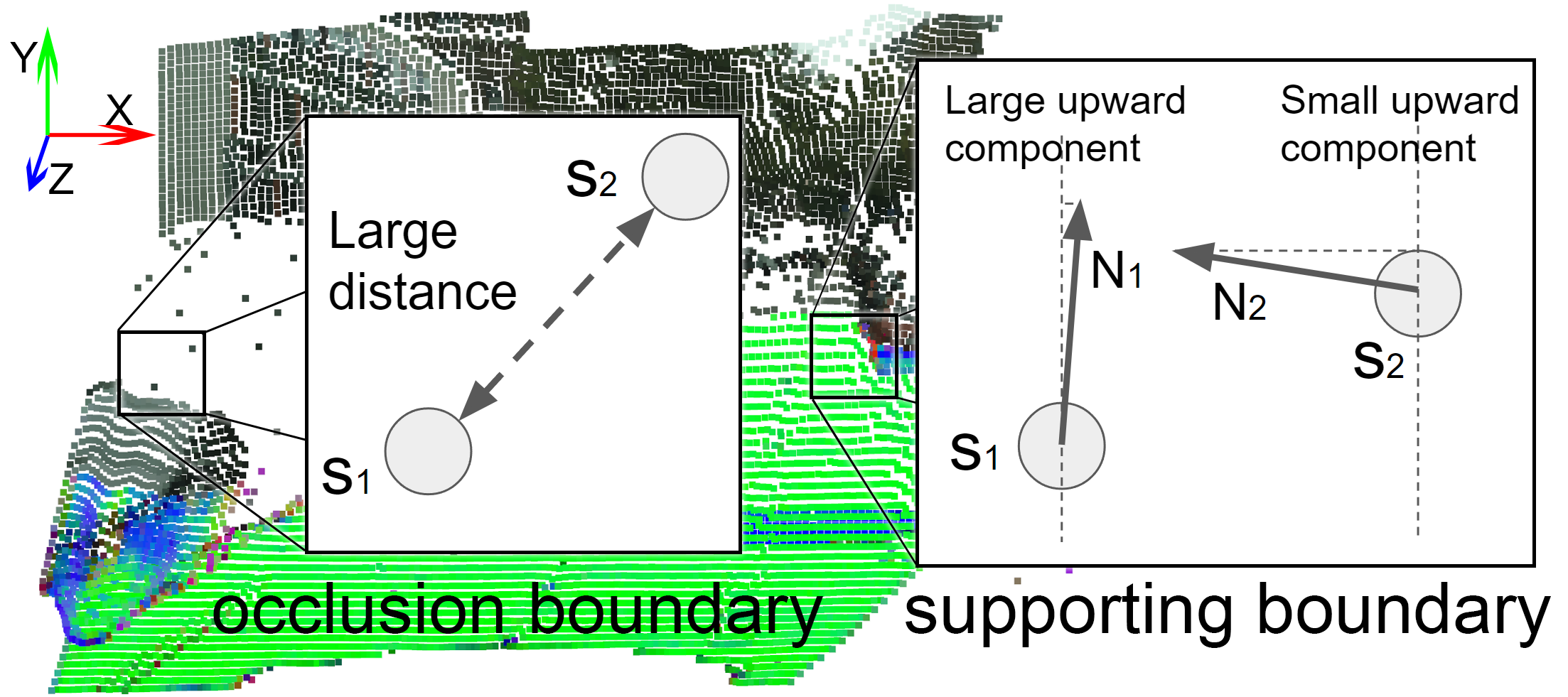}}\vspace{-0.2cm}\\

(a) &(b)\\
\bmvaHangBox{\includegraphics[width=5cm]{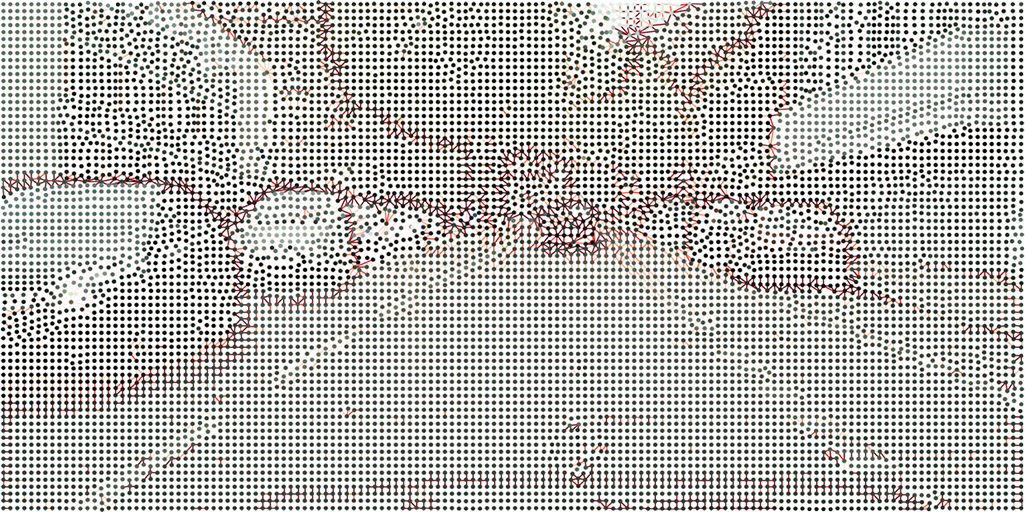}}&
\bmvaHangBox{\includegraphics[width=5cm]{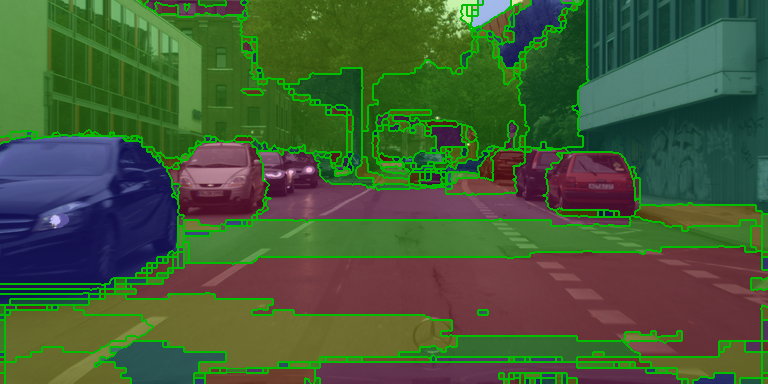}}\\
(c) &(d)\\
\end{tabular}
\caption{Decomposition of the proposed coherent pixels grouping algorithm based on depth. (a) Estimated depth and RGB image with SLIC superpixels. (b) Typical examples of occlusion and supporting boundaries. The bottom of point cloud is colored based on the normal vector. (c) Weighted boundary graph with weakly connected edges indicated in red. (d) The final grouping results from the community detection.}
\label{grouping}
\end{figure}

For each image, we compute the normal and 3D coordinate of pixels based on the depth $D$. We use SLIC superpixels~\cite{DBLP:journals/pami/AchantaSSLFS12} to downsample the image as shown in Figure~\ref{grouping} (a) and build a region adjacency graph from the superpixels. Each superpixel node contains the average 3d coordinate $[x,y,z]$ and upward normal $N_y$ perpendicular to the ground. With these attributes we weigh the edges based on how likely the edges lie on spatial boundaries.

We define two types of boundaries, namely occlusion and supporting boundaries~\cite{DBLP:conf/eccv/SilbermanHKF12}, as shown in Figure~\ref{grouping} (b). Let us consider two adjacent superpixels $\mathbf{S}_{1}$ and $\mathbf{S}_{2}$, where $\mathbf{S}_{1}$ has lower height $y_1$. To detect occlusion boundaries where the foreground occludes the background we compute the euclidean distance between $\mathbf{S}_{1}$ and $\mathbf{S}_{2}$ as in Eqn.~\ref{do}. Since the 3D scale of superpixels increases proportionally to depth, the distance should be normalized by their joint depth. To detect supporting boundaries where the objects rest on the ground, we assume that the lower $\mathbf{S}_{1}$ is the ground and $\mathbf{S}_{2}$ is an object. The pattern of urban-scene images implies that the ground surfaces should be level, which is measured by the normal in the upward direction $n_{y1}$. The ground should also be low which is measured by the square of $y_{1}$ when $y_1<0$. The objects on the ground should be upright, measured by the difference of their normals $n_{y1}$ and $n_{y2}$. The three terms are multiplied resulting in Eqn.~\ref{ds}.

\begin{equation}
\begin{aligned}
D_{ocln} = \frac{\sqrt{(x_1-x_2)^2+(y_1-y_2)^2+(z_1-z_2)^2}}{z_1+z_2}
\end{aligned}
\label{do}
\end{equation}

\begin{equation}
\begin{aligned}
D_{sup} = \left\{ \begin{aligned}\max(n_{y1},0)(\max(n_{y1} - n_{y2},0))y_{1}^2 & & {y_{1} < 0}\\
0 & & {y_{1} \geq 0}\end{aligned} \right.
\end{aligned}
\label{ds}
\end{equation}

Finally, the edge weight $W$ between $\mathbf{S}_1$ and $\mathbf{S}_2$ is computed as the weighted sum of the distance terms in Eqns.~\ref{do} and~\ref{ds} and a bias term, normalized by the sigmoid function, as shown in Eqn.~\ref{W12}.

\begin{equation}
\begin{aligned}
W = \sigma(w_{ocln}D_{ocln}+w_{sup}D_{sup}+b),
\end{aligned}
\label{W12}
\end{equation}
where $\sigma$ is the sigmoid function. With the weighted graph as shown in Figure~\ref{grouping} (c), our goal is to group strongly connected nodes while separating them from weakly connected nodes from other groups. Weak connections are indicated in red. As such we turn pixel grouping into a community detection problem~\cite{DBLP:journals/corr/abs-0906-0612,DBLP:conf/iwcia/BrowetAD11}. We use the classic InfoMap algorithm~\cite{Rosvall_2009} to fulfill the task. Details are provided in supplementary materials.

\subsection{Copy-Paste}

We copy-paste the coherent depth regions to build cross-context correspondences. We sample $M$ images and extract all regions above a certain size. Those images also act as backgrounds. We perform copy-paste in two rounds. All regions will be pasted $e$ times on every unmodified background in a round. In the first round, we set a small $e$ to display more parts of the original images. In the second round, $e$ is made large to allow more regions to be present in new contexts. The generated $2M$ images form a training sample. Details are provided in supplementary materials.

Along with copy-paste, random resize, horizontal flip, color jitter, and Gaussian blur are applied. To generate more realistic-looking images, we adopt DepthMix~\cite{DBLP:conf/cvpr/HoyerDCKSG21} to simulate occlusions. The depth of the copied patch and its target position are compared and only the pixels with smaller depth, i.e. closer to the camera are kept. Moreover, the height of the target position should be within a threshold $h_t$ of the original height of the patch. All geometric transformations are recorded as transformation matrices so that pixel correspondences can be traced back during training. 

\subsection{Contrastive Learning and Positive Samples Sampling}

\paragraph{SwAV contrastive learning framework}
We adopt the clustering-based contrastive learning framework introduced by SwAV~\cite{DBLP:conf/nips/CaronMMGBJ20} for its simplicity since it only relies on positive samples. We recap SwAV, with a slight modification in the swap prediction loss since the number of positive samples is inconsistent between each group. With a set of pixel representations denoted by $\mathbf{Z} = [\mathbf{z}_0,\mathbf{z}_1,...,\mathbf{z}_N]$ sampled from the feature maps, we compute their soft assignments $\mathbf{P} = [\mathbf{p}_0,\mathbf{p}_1,...,\mathbf{p}_N]$ to $K$ prototypes $\mathbf{C} = [\mathbf{c}_0,\mathbf{c}_1,...,\mathbf{c}_K]$ by Eqn.~\ref{P} with temperature $\tau$ as follows:
\begin{equation}
\mathbf{p}_{n}^{(k)}=\frac{\exp \left(\frac{1}{\tau} \mathbf{z}_{n}^{\top} \mathbf{c}_{k}\right)}{\sum_{k^{\prime}} \exp \left(\frac{1}{\tau} \mathbf{z}_{n}^{\top} \mathbf{c}_{k^{\prime}}\right)}
\label{P}
\end{equation}

Directly minimizing the cross-entropy for the assignments of positive samples will lead to a trivial solution. Codes $\mathbf{Q}$ are computed by Sinkhorn-Knopp algorithm~\cite{DBLP:conf/nips/Cuturi13} which is the balanced assignments with the smallest distance to $\mathbf{P}$. We compute the average code within each group of positive samples and get the target Codes $\overline{\mathbf{Q}} = [\overline{\mathbf{q}}_0,\overline{\mathbf{q}}_1,...,\overline{\mathbf{q}}_N]$ by Eqn.~\ref{Q}. 

\begin{equation}
\begin{aligned}
\overline{\mathbf{q}}_{n}^{(k)} = \frac{1}{\left| \mathbf{G}^{n}\right|} \sum_{i\sim \mathbf{G}^{n}} \mathbf{q}_{i}^{(k)} & & {n\sim \mathbf{G}^{n}}
\end{aligned}
\label{Q}
\end{equation}

The loss for $\mathbf{Z}$ is given by the mean of cross-entropy of all assignments and their corresponding codes as Eqn. ~\ref{Lz}.

\begin{equation}
\mathcal{L}_\mathbf{Z} =-\frac{1}{N} \sum_{n=1}^{N} \sum_{k=1}^{K} \overline{\mathbf{q}}_{n}^{(k)} \log \mathbf{p}_{n}^{(k)}
\label{Lz}
\end{equation}

\paragraph{Positive sampling}
The positive samples are defined in two ways. For pixel-level positive samples, the same pixels under different transformations are positive samples. We randomly sample a certain number of initial coordinates and locate their positive counterparts within the batch using the stored geometric transformations. For region-level positive samples, we assume the pixels inside the same coherent depth region share similarity. We randomly sample coordinates, where coordinates from the same region are considered positive samples. 

Both sampling methods can be used separately or jointly during training. Let $\mathcal{L}_{pixel}$ and $\mathcal{L}_{region}$ denote the loss computed among the pixel-level positive samples and region-level positive samples. We balance them with a weight $\lambda$ as in Eqn.~\ref{L}. 
\begin{equation}
\mathcal{L} =\lambda \mathcal{L}_{pixel} + (1-\lambda) \mathcal{L}_{region}
\label{L}
\end{equation}

\subsection{Clustering}
For unsupervised semantic segmentation, we need to cluster the learned representations. A fine-grained clustering into prototypes is already done by SwAV~\cite{DBLP:conf/nips/CaronMMGBJ20} and we can further group these to match the number of target classes. Empirically, we chose agglomerative clustering with cosine distance and average linkage criterion. 


\section{Experiments}
We evaluate on Cityscapes~\cite{DBLP:conf/cvpr/CordtsORREBFRS16} and KITTI~\cite{Geiger2013IJRR}. Cityscapes has 2975 30-frame stereo videos with one frame annotated in the $train$ set and 500 images in the $val$ set. KITTI has 71 stereo videos and 200 annotated images in the $train$ set. We only used the left images, resulting in 89250 images for Cityscapes and 42382 for KITTI. We resized Cityscapes images to $384\times768$ and KITTI images to $384\times1280$ pixels. We used semantic FPN~\cite{DBLP:conf/cvpr/KirillovGHD19} with ResNet-50~\cite{He_2016_CVPR} as the feature extractor. The final classifier was replaced by a 1-layer MLP projection head with $128$ output channels. All experiments are conducted on a single 16GB V100 GPU. Further details on training and evaluation setting can be found in supplementary materials.

\subsection{Unsupervised Semantic Segmentation}
\begin{figure}[H]
\setlength\tabcolsep{2pt}
\centering
\begin{tabular}{ccccccc}
\centering
\bmvaHangBox{\includegraphics[width=1.4cm]{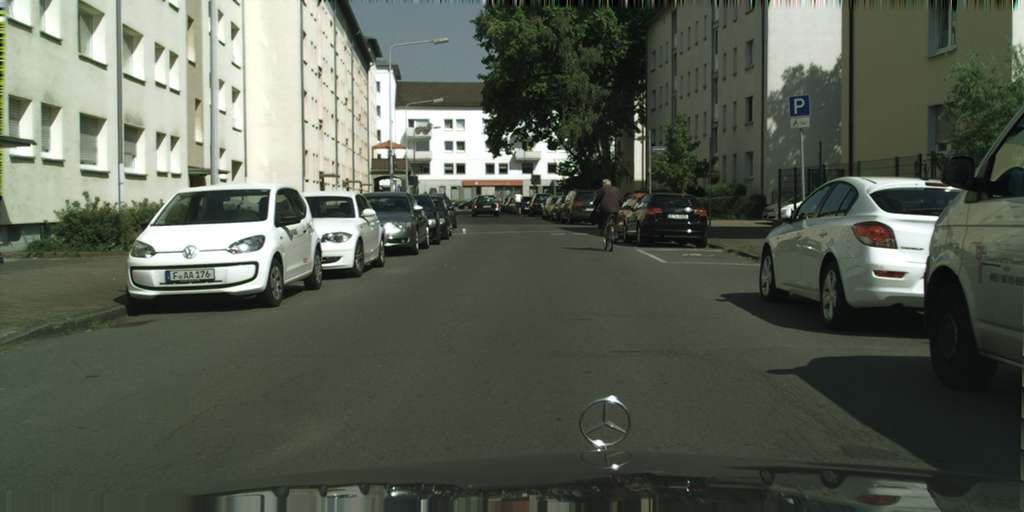}}&\bmvaHangBox{\includegraphics[width=1.4cm]{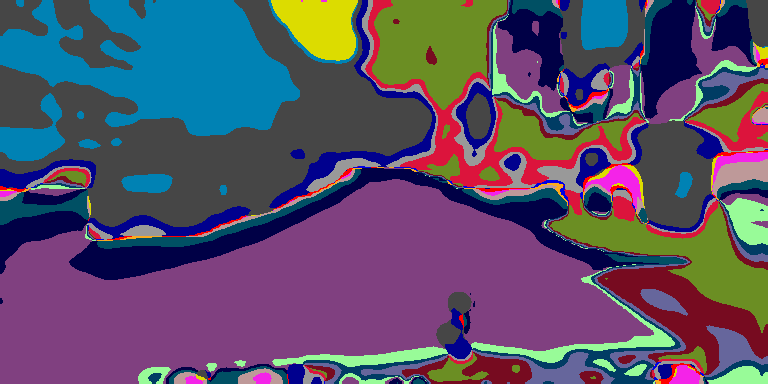}}&\bmvaHangBox{\includegraphics[width=1.4cm]{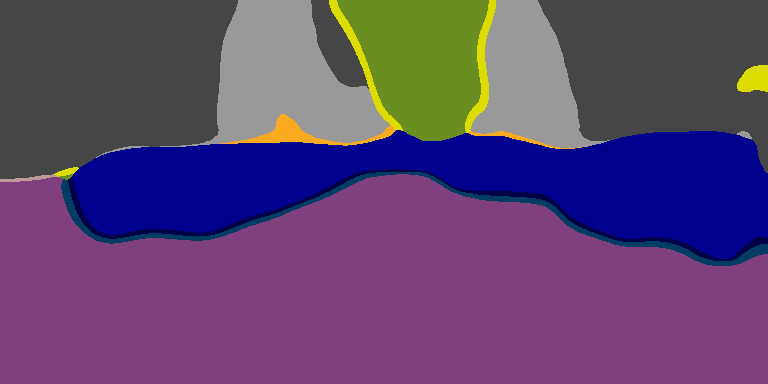}}&\bmvaHangBox{\includegraphics[width=1.4cm]{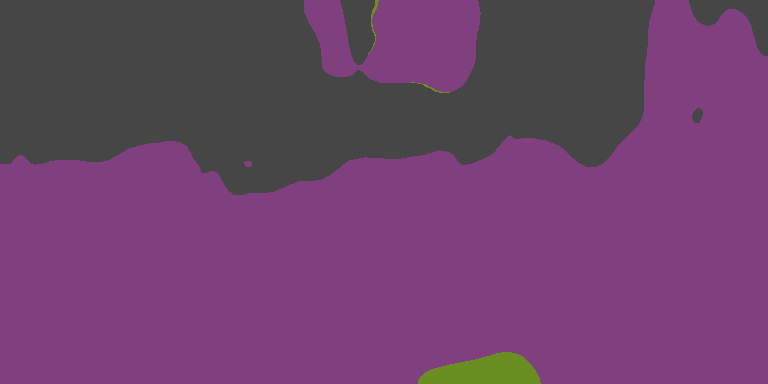}}&\bmvaHangBox{\includegraphics[width=1.4cm]{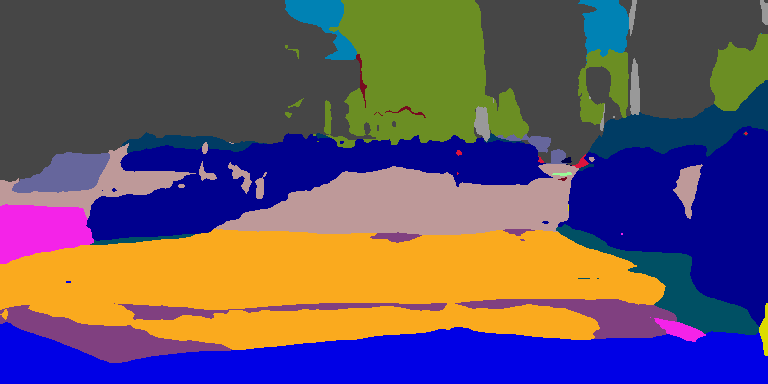}}&\bmvaHangBox{\includegraphics[width=1.4cm]{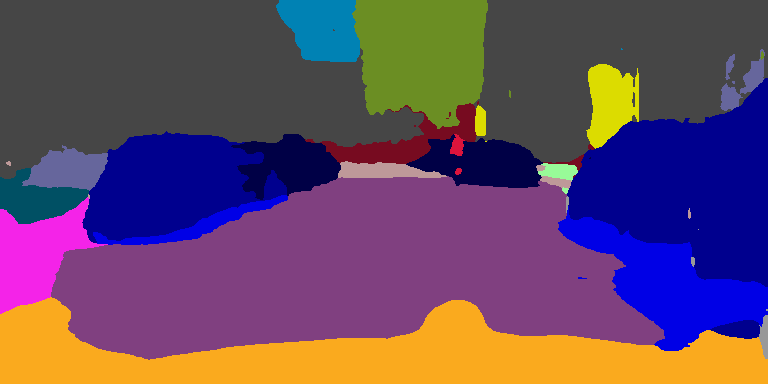}}&\bmvaHangBox{\includegraphics[width=1.4cm]{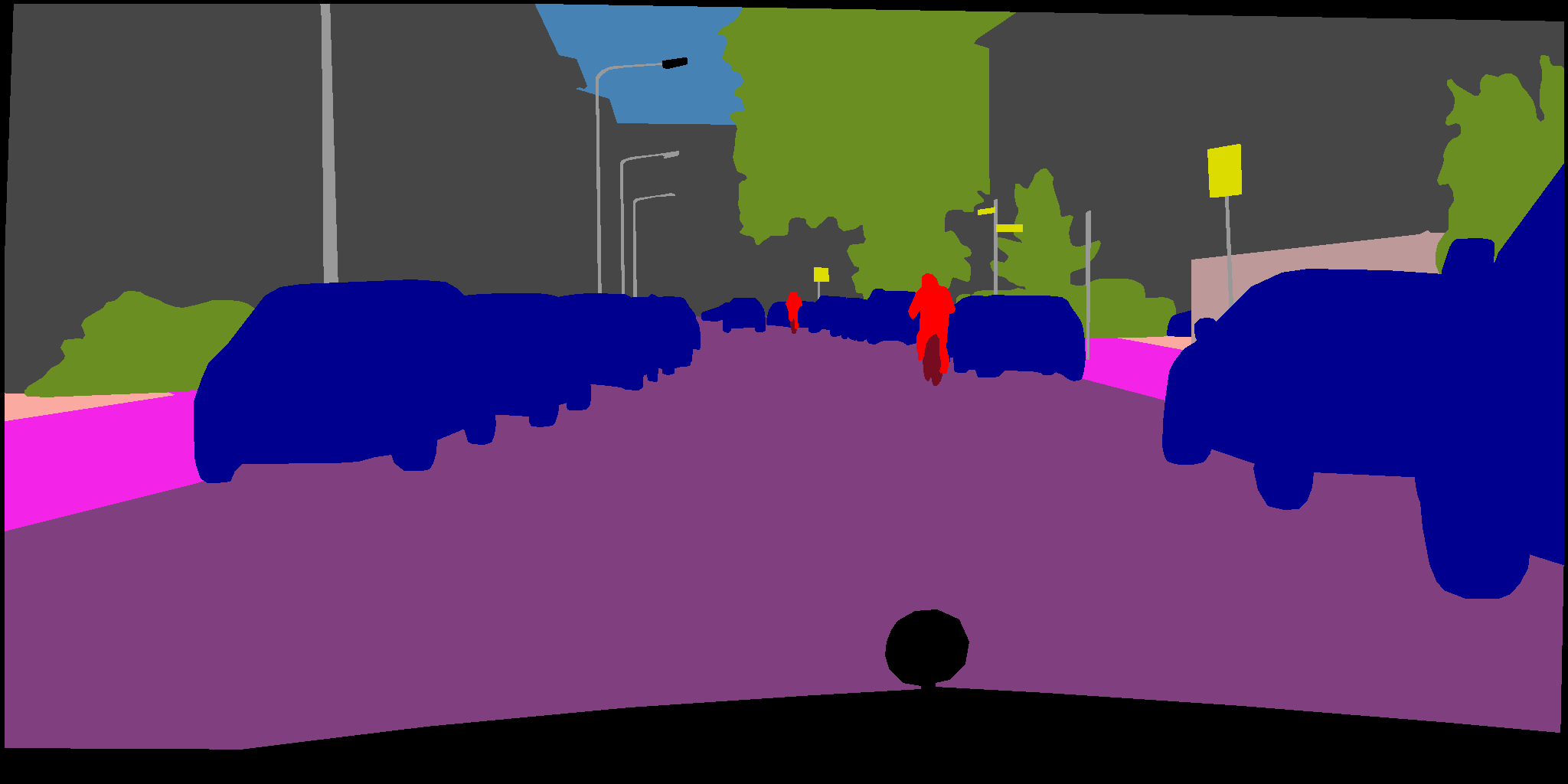}}
\\
\bmvaHangBox{\includegraphics[width=1.4cm]{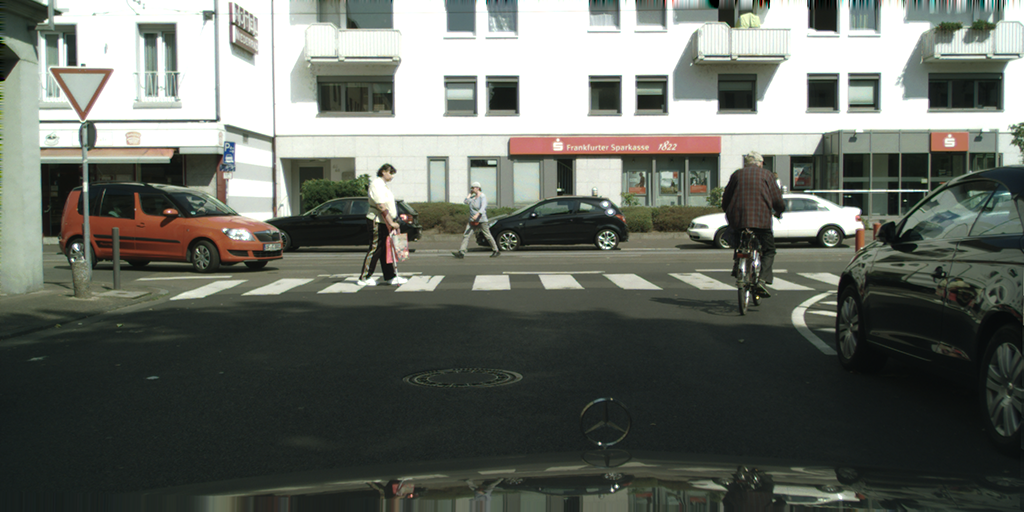}}&\bmvaHangBox{\includegraphics[width=1.4cm]{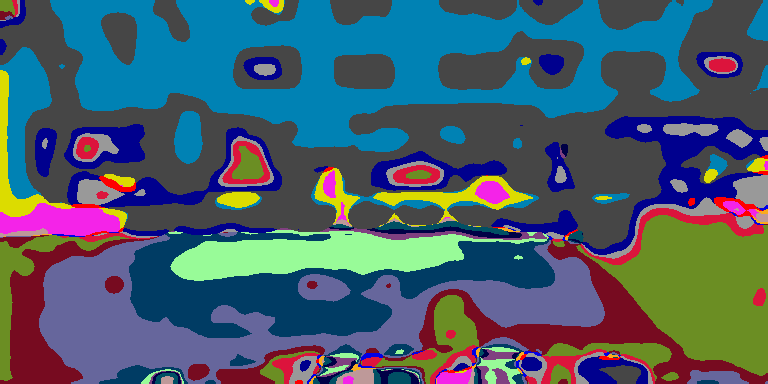}}&\bmvaHangBox{\includegraphics[width=1.4cm]{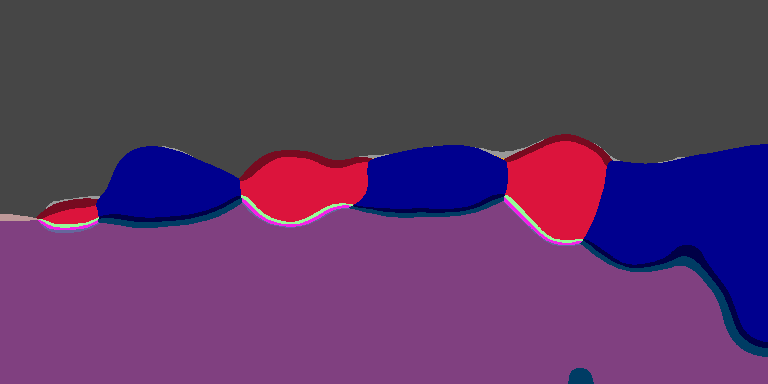}}&\bmvaHangBox{\includegraphics[width=1.4cm]{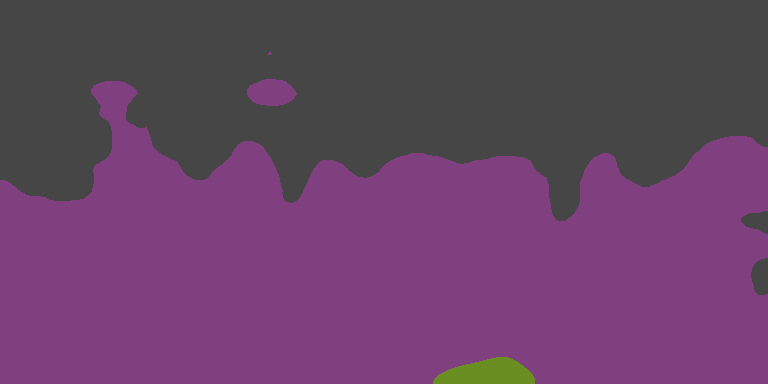}}&\bmvaHangBox{\includegraphics[width=1.4cm]{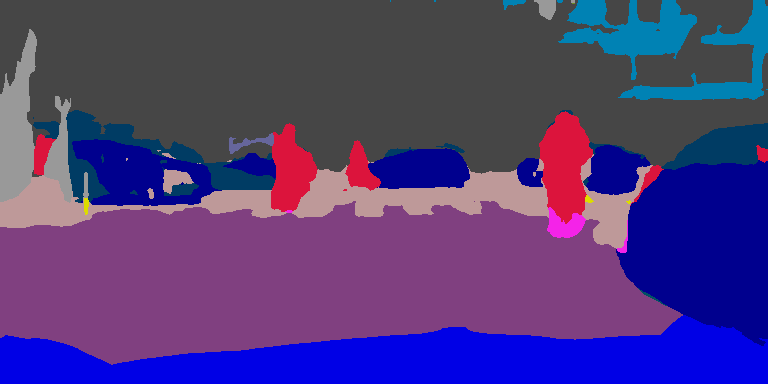}}&\bmvaHangBox{\includegraphics[width=1.4cm]{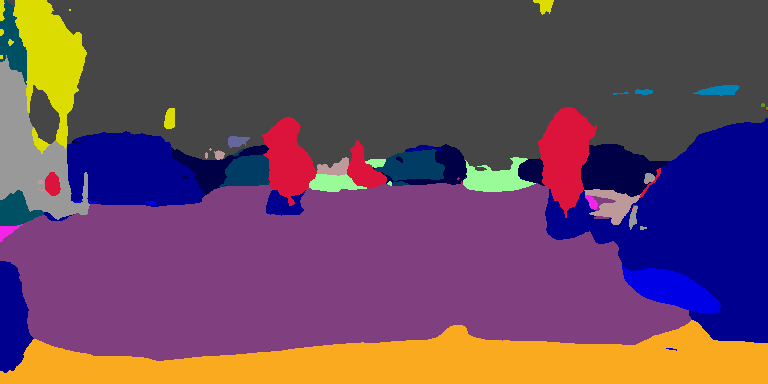}}&\bmvaHangBox{\includegraphics[width=1.4cm]{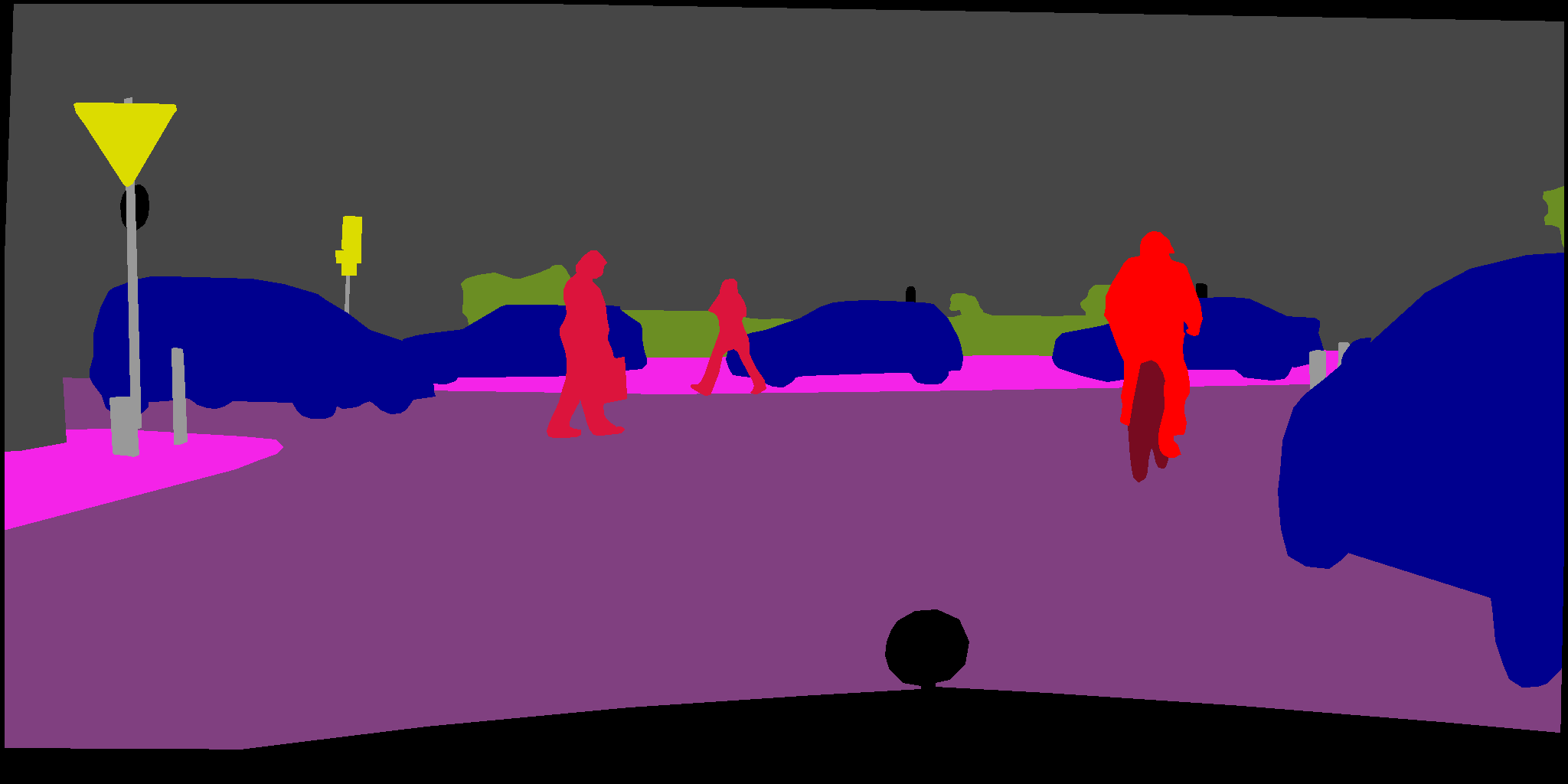}}\\
Image&PiCIE&PiCIE-IN&PiCIE-IN& Ours & Ours-IN & GT\\
&&5 epochs&20 epochs&&
\end{tabular}
\caption{Qualitative comparison of unsupervised semantic segmentation on Cityscapes.}
\label{unsupervisedquality}
\end{figure}

Segmentation can be generated by clustering over the learned dense visual representation directly. The unsupervised semantic segmentation performance of our method and PiCIE~\cite{DBLP:conf/cvpr/ChoMBH21} with ImageNet pre-training or random initialization is shown in Tab.~\ref{us}. Using ImageNet initialization~\cite{ILSVRC15}, our method surpasses PiCIE~\cite{DBLP:conf/cvpr/ChoMBH21} by $+7.14\%$ and $6.65\%$ in mIoU on Cityscapes~\cite{DBLP:conf/cvpr/CordtsORREBFRS16} and KITTI~\cite{Geiger2013IJRR}, respectively. Our method can better capture the outline of objects, as shown in Figure~\ref{unsupervisedquality}, whereas PiCIE can only predict the coarse masks of big objects and \textit{stuff}, resulting in a good accuracy but poor mIoU. Our method performs more consistently when training on Cityscapes~\cite{DBLP:conf/cvpr/CordtsORREBFRS16} then testing on KITTI~\cite{Geiger2013IJRR}, or vice versa. We also observe that PiCIE~\cite{DBLP:conf/cvpr/ChoMBH21} performs poorly when initialized from scratch and diverges to a trivial solution as training progresses. Our method does not suffer from these shortcomings and is able to learn semantic clustering also when initialized from scratch, outperforming PiCIE~\cite{DBLP:conf/cvpr/ChoMBH21} by a large margin in both accuracy and mIoU.

\begin{table}[H]
  \centering
  \begin{tabular}{@{}lllcccc@{}}
    \toprule
    Method & Init. & Training Data &  \multicolumn{2}{c}{CS-Sem.} & \multicolumn{2}{c}{KT-Sem.} \\
     &  &  &  $Acc$ & $mIoU$ &  $Acc$ & $mIoU$ \\
    \midrule
    PiCIE~\cite{DBLP:conf/cvpr/ChoMBH21} & scratch & CS~\cite{DBLP:conf/cvpr/CordtsORREBFRS16} & 33.56 & 8.33 & 32.20 & 6.52\\
    Ours($\lambda=0.5$) & scratch & CS~\cite{DBLP:conf/cvpr/CordtsORREBFRS16} & 65.42 & 20.49 & \textbf{68.37} & 21.03\\
    PiCIE~\cite{DBLP:conf/cvpr/ChoMBH21} & scratch & KT~\cite{DBLP:conf/cvpr/CordtsORREBFRS16} & 30.28 & 6.81 & 30.62 & 7.54\\
    Ours($\lambda=0.5$) & scratch & KT~\cite{Geiger2013IJRR} & 49.18 & 17.20 & 49.58 & 18.22\\
    PiCIE~\cite{DBLP:conf/cvpr/ChoMBH21} & IN~\cite{ILSVRC15} & CS~\cite{DBLP:conf/cvpr/CordtsORREBFRS16} & \textbf{68.50} & 16.24 & 56.74 & 13.54\\
    Ours($\lambda=0.5$) & IN~\cite{ILSVRC15} & CS~\cite{DBLP:conf/cvpr/CordtsORREBFRS16} & 66.70 & \textbf{23.38} & 68.25 & \textbf{22.50}\\
    PiCIE~\cite{DBLP:conf/cvpr/ChoMBH21} & IN~\cite{ILSVRC15} & KT~\cite{DBLP:conf/cvpr/CordtsORREBFRS16} & 53.24 & 12.55 & 61.74 & 12.92 \\
    Ours($\lambda=0.5$) & IN~\cite{ILSVRC15} & KT~\cite{Geiger2013IJRR} & 56.96 & 18.85 & 59.11 & 19.57\\
    \bottomrule
  \end{tabular}
  \caption{Unsupervised semantic segmentation performance on Cityscapes~\cite{DBLP:conf/cvpr/CordtsORREBFRS16} $val$ set and KITTI~\cite{Geiger2013IJRR} $train$ set. We retrained PiCIE with equivalent setting to ours. IN: ImageNet~\cite{ILSVRC15}; CS: Cityscapes~\cite{DBLP:conf/cvpr/CordtsORREBFRS16}; KT: KITTI~\cite{Geiger2013IJRR}.}
  \label{us}
\end{table}

\vspace{-0.6cm}
\subsection{Semi-Supervised Segmentation}
In semi-supervised segmentation the self-supervised pre-trained models are used as initialization for the following supervised learning. For urban scene segmentation a common practice is to pre-train on large scale datasets, such as ImageNet~\cite{ILSVRC15} and COCO~\cite{DBLP:conf/eccv/LinMBHPRDZ14}. Our method, in contrast, allows directly pre-training on the target urban scene data, without the need for extra curated datasets. 

We compared our pre-trained model with existing models~\cite{DBLP:conf/nips/CaronMMGBJ20,DBLP:conf/cvpr/WangZSKL21,DBLP:conf/cvpr/XieL00L021,xie2021unsupervised,DBLP:conf/cvpr/SelvarajuD0N21} trained on ImageNet~\cite{ILSVRC15} and COCO~\cite{DBLP:conf/eccv/LinMBHPRDZ14} by their original authors. However, pre-trained models on Cityscapes~\cite{DBLP:conf/cvpr/CordtsORREBFRS16} and KITTI~\cite{Geiger2013IJRR} are not publicly available for the baseline methods and neither are we able to train them using the original setting due to limited compute power. We trained SwAV~\cite{DBLP:conf/nips/CaronMMGBJ20} and PixPro~\cite{DBLP:conf/cvpr/XieL00L021} on Cityscapes~\cite{DBLP:conf/cvpr/CordtsORREBFRS16} or KITTI~\cite{Geiger2013IJRR} on a single 16 GB GPU using their official implementations. In this experiment, all models were initialized from scratch in pre-training and used ResNet-50~\cite{He_2016_CVPR} backbone.

\begin{table}[H]
  \centering
  \begin{tabular}{@{}llccccccc@{}}
    \toprule
    Pre-training &Pre-training  & CS-Sem. & \multicolumn{2}{c}{CS-Inst.}& KT-Sem. & \multicolumn{2}{c}{KT-Inst.}\\
    Method & Dataset & $mIoU$ & $AP$ & $AP_{50}$& $mIoU$ & $AP$ & $AP_{50}$ \\
    \midrule
    scratch & - & 65.11 & 24.30 & 46.98 & 32.99 & 8.15 & 15.79\\
    supervised & IN~\cite{ILSVRC15} & 70.54 & 27.34 & 50.59 & 40.09 & 12.38 & 23.42\\
    SwAV~\cite{DBLP:conf/nips/CaronMMGBJ20} & IN~\cite{ILSVRC15} & 71.07 & 28.08 & 52.25 & 40.52 & \textbf{13.78} & \textbf{27.90} \\
    DenseCL~\cite{DBLP:conf/cvpr/WangZSKL21} & IN~\cite{ILSVRC15} & 72.09 & 28.97 & 51.93 & 40.88 & 12.63 & 22.74\\
    PixPro~\cite{DBLP:conf/cvpr/XieL00L021} & IN~\cite{ILSVRC15} & 72.66 & 29.04 & 52.59 & 40.50 & 13.04 & 24.95\\
    ORL~\cite{xie2021unsupervised} & CC~\cite{DBLP:conf/eccv/LinMBHPRDZ14} & 72.32 & \textbf{29.94} & 52.55 & 41.88 & 12.02 & 23.48\\
    CAST~\cite{DBLP:conf/cvpr/SelvarajuD0N21} & CC~\cite{DBLP:conf/eccv/LinMBHPRDZ14} & 69.92 & 27.33 & 51.31 & 38.78 & 10.67 & 20.13\\
    SwAV~\cite{DBLP:conf/nips/CaronMMGBJ20} & CS~\cite{DBLP:conf/cvpr/CordtsORREBFRS16} & 61.69 & 23.62& 46.21 & 36.10 & 9.22 & 18.01\\
    PixPro~\cite{DBLP:conf/cvpr/XieL00L021} & CS~\cite{DBLP:conf/cvpr/CordtsORREBFRS16} & 61.64 & 23.78 & 46.45 & 36.99 & 9.61 & 18.73 \\
    SwAV~\cite{DBLP:conf/nips/CaronMMGBJ20} & KT~\cite{DBLP:conf/cvpr/CordtsORREBFRS16} & 60.74 & 23.51 & 46.08 & 36.90 & 9.42 & 18.11\\
    PixPro~\cite{DBLP:conf/cvpr/XieL00L021} & KT~\cite{DBLP:conf/cvpr/CordtsORREBFRS16} & 61.25 & 23.23 & 46.23 & 37.28 & 9.45 & 18.57\\
    Ours($\lambda=1$) & CS~\cite{DBLP:conf/cvpr/CordtsORREBFRS16} & \textbf{73.55} & \textbf{29.94} & \textbf{52.88} & \textbf{42.70} & 12.58 & 24.98\\
    Ours($\lambda=0.5$) & CS~\cite{DBLP:conf/cvpr/CordtsORREBFRS16} & 73.03 & 29.11 & 51.87 & 42.32 & 12.22 & 23.16\\
    Ours($\lambda=1$) & KT~\cite{Geiger2013IJRR} & 71.62 & 28.77 & 52.71 & 41.17 & 11.74 &  20.57\\
    Ours($\lambda=0.5$) & KT~\cite{Geiger2013IJRR} & 71.45 & 27.86 & 51.16 & 41.03 & 11.36 &  20.49\\
    \bottomrule
  \end{tabular}
  \caption{Segmentation performance over Cityscapes $val$ set and 5-fold validation of KITTI $train$ set. SwAV and PixPro on Cityscapes and KITTI are trained with limited GPU compute as ours. IN: ImageNet~\cite{ILSVRC15}; CS: Cityscapes~\cite{DBLP:conf/cvpr/CordtsORREBFRS16}; CC: COCO~\cite{DBLP:conf/eccv/LinMBHPRDZ14}; KT: KITTI~\cite{Geiger2013IJRR}.}
  \label{ft}
\end{table}

\paragraph{Semantic segmentation} We show semi-supervised semantic segmentation performance on Cityscapes~\cite{DBLP:conf/cvpr/CordtsORREBFRS16} and KITTI~\cite{Geiger2013IJRR} in Tab.~\ref{ft}. Our method pre-trained on Cityscapes~\cite{DBLP:conf/cvpr/CordtsORREBFRS16} with $\lambda=1$ outperforms other existing pre-trained models for semantic segmentation by a noticeable margin on both datasets. It exceeds the SwAV~\cite{DBLP:conf/nips/CaronMMGBJ20} pre-trained model by $+3.01\%$ on Cityscapes~\cite{DBLP:conf/cvpr/CordtsORREBFRS16} and $+2.18\%$ on KITTI~\cite{Geiger2013IJRR} in ${\rm mIoU}$. The KITTI pre-trained model also achieves a comparable result on both datasets with less training data. Overall, our method demonstrates better performance and generalization across different urban scene datasets.
\paragraph{Instance segmentation} We show semi-supervised instance segmentation performance in Tab.~\ref{ft} on Cityscapes~\cite{DBLP:conf/cvpr/CordtsORREBFRS16} and KITTI~\cite{Geiger2013IJRR}. Our Cityscapes pre-trained model with $\lambda=1$ still slightly outperforms other baselines on Cityscapes~\cite{DBLP:conf/cvpr/CordtsORREBFRS16}. It still exceeds SwAV~\cite{DBLP:conf/nips/CaronMMGBJ20} by $+1.86\%$ in ${\rm AP}$ and $+0.63\%$ in ${\rm AP_{50}}$ on Cityscapes~\cite{DBLP:conf/cvpr/CordtsORREBFRS16}. The performance on KITTI~\cite{DBLP:conf/cvpr/CordtsORREBFRS16} is below SwAV~\cite{DBLP:conf/nips/CaronMMGBJ20} by $-1.20\%$ in ${\rm AP}$ and $-2.92\%$ in ${\rm AP_{50}}$. This is expected as some layers of semantic FPN~\cite{DBLP:conf/cvpr/KirillovGHD19} are discarded to adapt the Mask-RCNN~\cite{DBLP:conf/iccv/HeGDG17} architecture for instance segmentation.

\paragraph{Training baseline methods on urban scenes}  Pre-training on Cityscapes and KITTI with SwAV~\cite{DBLP:conf/nips/CaronMMGBJ20} and PixPro~\cite{DBLP:conf/cvpr/XieL00L021} yields inferior results. It indicates that the improved performance of our method is not only due to pre-training on similar data to the downstream tasks, but also from our effective design, especially under limited compute. Training details and discussion are provided in the supplementary material.

\subsection{Ablation Studies}

We performed ablation studies to validate our design choices. In ablation experiments, the models were trained on the 2975 images from the $train$ set of the Cityscapes~\cite{DBLP:conf/cvpr/CordtsORREBFRS16} for 100 epochs. The semantic segmentation performance is reported after fine-tuning on $1/16$ of the $train$ set for 6k iterations. 

\begin{figure}[H]
\centering
\begin{tabular}{ccc}
\centering
\bmvaHangBox{\includegraphics[width=3.5cm]{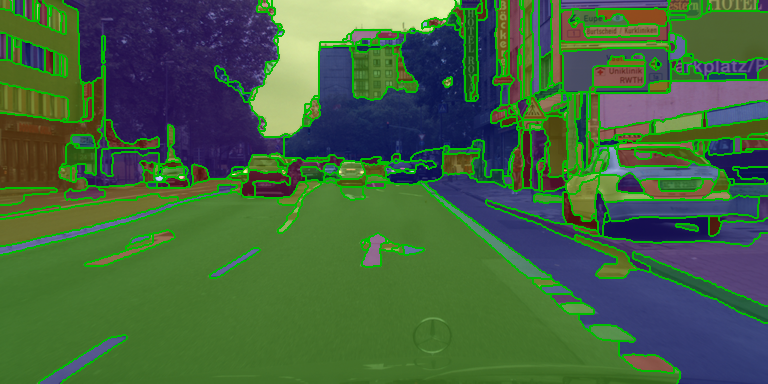}}&
\bmvaHangBox{\includegraphics[width=3.5cm]{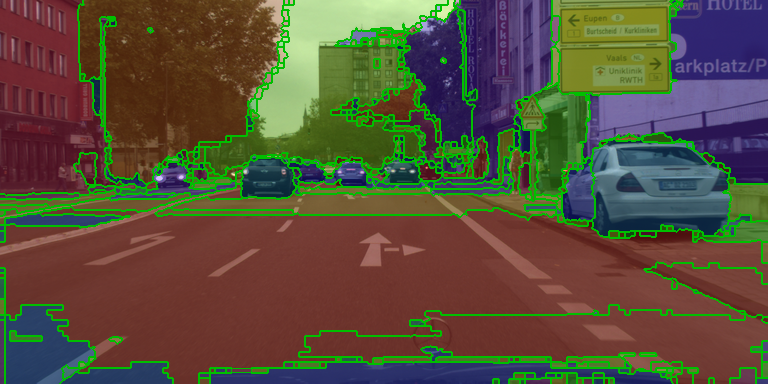}}&
\bmvaHangBox{\includegraphics[width=3.5cm]{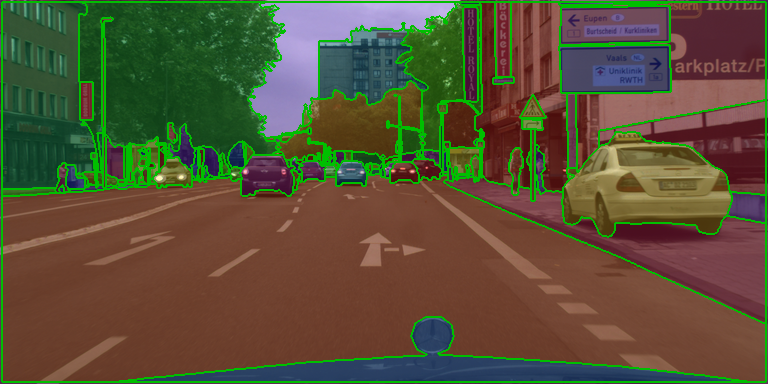}}\\
(a)owt-ucm~\cite{DBLP:journals/pami/ArbelaezMFM11}&(b)Ours&(c)GT
\end{tabular}
\caption{Examples of different region proposals. }
\label{proposal}
\end{figure}

\paragraph{Region proposal} To study the significance of our coherent depth region proposal we compared different region proposal methods, including owt-ucm~\cite{DBLP:journals/pami/ArbelaezMFM11} used by Zhang et al.~\cite{DBLP:conf/nips/ZhangM20}, as shown in Figure~\ref{proposal}. We also split the instance label in Cityscapes~\cite{DBLP:conf/cvpr/CordtsORREBFRS16} $train$ into only connected regions as the optimal region proposals (ground truth). We show the fine-tuning performance in Tab.~\ref{ablation2}. Our coherent depth region proposal produces the closest performance to the ground truth.

\begin{table}[H]
\centering
\begin{tabular}{c|ccc|ccc}
\toprule
 & \multicolumn{3}{c}{Fine-tuning}\vline&\multicolumn{3}{c}{Clustering}\\
Weight&$\lambda = 0$&$\lambda = 0.5$&$\lambda = 1$&$\lambda = 0$&$\lambda = 0.5$&$\lambda = 1$ \\
\midrule
owt-ucm~\cite{DBLP:journals/pami/ArbelaezMFM11}&44.83&47.01&47.24&13.64&15.70&12.11\\
Ours&46.91&48.87&48.55&15.94&20.71&13.94\\
GT&49.92&52.26&48.82&24.05&27.35&14.59\\

\bottomrule
\end{tabular}
\caption{Effects of different types of object proposals. Measurement is based on mIoU by fine-tuning on a subset of Cityscapes~\cite{DBLP:conf/cvpr/CordtsORREBFRS16} for semantic segmentation.}
\label{ablation1}
\end{table}

\paragraph{Copy-Paste} To further investigate the effectiveness of copy-paste, we altered the number of images $M$ for copy-paste. We report the fine-tuning and clustering performance in Tab.~\ref{ablation2}. Mixing more images to build richer cross-context correspondence is beneficial 
in all cases, especially when region-level positive samples are sampled ($\lambda<1$). 

\begin{table}[H]
\centering
\begin{tabular}{@{}ll|cccc|cccc@{}}
\toprule
\multicolumn{2}{c}{M} \vline & 0 & 2 & 4 & 8 & 0 & 2 & 4 & 8 \\
Weight& Proposal & \multicolumn{4}{c}{Fine-tuning} \vline& \multicolumn{4}{c}{Clustering}\\
\midrule
\multirow{2}*{$\lambda = 0$}
&Ours&35.18&42.34&46.75&46.91&5.93&13.92&13.65&15.94\\
&GT&37.70&46.74&48.64&49.92&9.39&21.31&23.51&24.05\\
\midrule
\multirow{2}*{$\lambda = 0.5$}
&Ours&34.35&45.92&48.23&48.87&3.97&14.03&20.32&20.71\\
&GT&38.01&49.26&50.06&52.26&8.20&24.26&24.82&27.35\\
\midrule
\multirow{2}*{$\lambda = 1$}
&Ours&\multirow{2}*{39.93}&48.24&48.09&48.55&\multirow{2}*{4.89}&10.59&13.65&13.94\\
&GT&&48.20&49.25&48.81&&10.36&14.08&14.59\\
\bottomrule
\end{tabular}
\caption{Effects of loss weighting (Eqn. \ref{L}) and number of sample images $M$ for copy-paste. 0 means copy-paste is disabled. Measurement is based on mIoU by fine-tuning on subset of Cityscapes~\cite{DBLP:conf/cvpr/CordtsORREBFRS16} for semantic segmentation.}
\label{ablation2}
\end{table}

\paragraph{Loss Weight} The effects of the loss weight in Eqn.~\ref{L} are shown in Tab.~\ref{ablation2}. With high-quality proposals like the ground truth, combining the pixel-level and region-level positive samples remarkably increases the performance for both fine-tuning and unsupervised clustering. Our coherent depth region proposals fail to strictly contain a single class as Figure ~\ref{proposal}b. Thus, including region-level positive samples takes little effect on fine-tuning performance as shown in Tab.~\ref{ablation2} or even slightly degrades the performance as shown in Tab.~\ref{ft}. However, for unsupervised segmentation performance, combining both pixel and region level positive samples is still beneficial.

\section{Discussion and Limitations}

In this work we leverage estimated depth to group semantically related coherent pixels, which allows copy-paste to build cross-context correspondences for contrastive learning. Experiments on Cityscapes~\cite{DBLP:conf/cvpr/CordtsORREBFRS16} and KITTI~\cite{Geiger2013IJRR} show that our method trained with one GPU surpasses previous state-of-the-art on unsupervised semantic segmentation and achieves similar fine-tuning performance to existing methods that are pre-trained on ImageNet~\cite{ILSVRC15} or COCO~\cite{DBLP:conf/eccv/LinMBHPRDZ14} with 8 GPUs. Our experiments also demonstrate the importance of using high quality region proposals for our method. The current pixel grouping algorithm is handcrafted and requires some task- and dataset-dependent parameter tuning. Future work therefore includes investigating data-driven approaches. Lastly, our method is particularly well suited to urban scene related datasets but its not clear if it is also applicable in other scenarios where depth is available or can be extracted. One inherent limitation of relying on self-supervised depth estimation is that these methods are developed and evaluated almost exclusively on urban scene images. Further experiments are needed to evaluate how reliably depth can be extracted in e.g. indoor scenes, and how well our method generalizes in these new settings. We hope our work inspires community to explore new applications and uses of depth in self-supervised representation learning. We invite researchers to explore and build upon our publicly available codebase.

\bibliography{egbib}

\begin{thebibliography}{59}
\providecommand{\natexlab}[1]{#1}
\providecommand{\url}[1]{\texttt{#1}}
\expandafter\ifx\csname urlstyle\endcsname\relax
  \providecommand{\doi}[1]{doi: #1}\else
  \providecommand{\doi}{doi: \begingroup \urlstyle{rm}\Url}\fi

\bibitem[Achanta et~al.(2012)Achanta, Shaji, Smith, Lucchi, Fua, and
  S{\"{u}}sstrunk]{DBLP:journals/pami/AchantaSSLFS12}
Radhakrishna Achanta, Appu Shaji, Kevin Smith, Aur{\'{e}}lien Lucchi, Pascal
  Fua, and Sabine S{\"{u}}sstrunk.
\newblock {SLIC} superpixels compared to state-of-the-art superpixel methods.
\newblock \emph{{IEEE} Trans. Pattern Anal. Mach. Intell.}, 34\penalty0
  (11):\penalty0 2274--2282, 2012.
\newblock \doi{10.1109/TPAMI.2012.120}.
\newblock URL \url{https://doi.org/10.1109/TPAMI.2012.120}.

\bibitem[Arbelaez et~al.(2011)Arbelaez, Maire, Fowlkes, and
  Malik]{DBLP:journals/pami/ArbelaezMFM11}
Pablo Arbelaez, Michael Maire, Charless~C. Fowlkes, and Jitendra Malik.
\newblock Contour detection and hierarchical image segmentation.
\newblock \emph{{IEEE} Trans. Pattern Anal. Mach. Intell.}, 33\penalty0
  (5):\penalty0 898--916, 2011.
\newblock \doi{10.1109/TPAMI.2010.161}.
\newblock URL \url{https://doi.org/10.1109/TPAMI.2010.161}.

\bibitem[Barnea and Ben{-}Shahar(2019)]{DBLP:conf/cvpr/BarneaB19}
Ehud Barnea and Ohad Ben{-}Shahar.
\newblock Exploring the bounds of the utility of context for object detection.
\newblock In \emph{{IEEE} Conference on Computer Vision and Pattern
  Recognition, {CVPR} 2019, Long Beach, CA, USA, June 16-20, 2019}, pages
  7412--7420. Computer Vision Foundation / {IEEE}, 2019.
\newblock \doi{10.1109/CVPR.2019.00759}.
\newblock URL
  \url{http://openaccess.thecvf.com/content\_CVPR\_2019/html/Barnea\_Exploring\_the\_Bounds\_of\_the\_Utility\_of\_Context\_for\_Object\_CVPR\_2019\_paper.html}.

\bibitem[Browet et~al.(2011)Browet, Absil, and
  Dooren]{DBLP:conf/iwcia/BrowetAD11}
Arnaud Browet, Pierre{-}Antoine Absil, and Paul~Van Dooren.
\newblock Community detection for hierarchical image segmentation.
\newblock In Jake~K. Aggarwal, Reneta~P. Barneva, Valentin~E. Brimkov, Kostadin
  Koroutchev, and Elka Korutcheva, editors, \emph{Combinatorial Image Analysis
  - 14th International Workshop, {IWCIA} 2011, Madrid, Spain, May 23-25, 2011.
  Proceedings}, volume 6636 of \emph{Lecture Notes in Computer Science}, pages
  358--371. Springer, 2011.
\newblock \doi{10.1007/978-3-642-21073-0\_32}.
\newblock URL \url{https://doi.org/10.1007/978-3-642-21073-0\_32}.

\bibitem[Caron et~al.(2020)Caron, Misra, Mairal, Goyal, Bojanowski, and
  Joulin]{DBLP:conf/nips/CaronMMGBJ20}
Mathilde Caron, Ishan Misra, Julien Mairal, Priya Goyal, Piotr Bojanowski, and
  Armand Joulin.
\newblock Unsupervised learning of visual features by contrasting cluster
  assignments.
\newblock In Hugo Larochelle, Marc'Aurelio Ranzato, Raia Hadsell,
  Maria{-}Florina Balcan, and Hsuan{-}Tien Lin, editors, \emph{Advances in
  Neural Information Processing Systems 33: Annual Conference on Neural
  Information Processing Systems 2020, NeurIPS 2020, December 6-12, 2020,
  virtual}, 2020.
\newblock URL
  \url{https://proceedings.neurips.cc/paper/2020/hash/70feb62b69f16e0238f741fab228fec2-Abstract.html}.

\bibitem[Chen et~al.(2020{\natexlab{a}})Chen, Kornblith, Norouzi, and
  Hinton]{DBLP:conf/icml/ChenK0H20}
Ting Chen, Simon Kornblith, Mohammad Norouzi, and Geoffrey~E. Hinton.
\newblock A simple framework for contrastive learning of visual
  representations.
\newblock In \emph{Proceedings of the 37th International Conference on Machine
  Learning, {ICML} 2020, 13-18 July 2020, Virtual Event}, volume 119 of
  \emph{Proceedings of Machine Learning Research}, pages 1597--1607. {PMLR},
  2020{\natexlab{a}}.
\newblock URL \url{http://proceedings.mlr.press/v119/chen20j.html}.

\bibitem[Chen et~al.(2020{\natexlab{b}})Chen, Kornblith, Swersky, Norouzi, and
  Hinton]{DBLP:conf/nips/ChenKSNH20}
Ting Chen, Simon Kornblith, Kevin Swersky, Mohammad Norouzi, and Geoffrey~E.
  Hinton.
\newblock Big self-supervised models are strong semi-supervised learners.
\newblock In Hugo Larochelle, Marc'Aurelio Ranzato, Raia Hadsell,
  Maria{-}Florina Balcan, and Hsuan{-}Tien Lin, editors, \emph{Advances in
  Neural Information Processing Systems 33: Annual Conference on Neural
  Information Processing Systems 2020, NeurIPS 2020, December 6-12, 2020,
  virtual}, 2020{\natexlab{b}}.
\newblock URL
  \url{https://proceedings.neurips.cc/paper/2020/hash/fcbc95ccdd551da181207c0c1400c655-Abstract.html}.

\bibitem[Chen and He(2021)]{DBLP:conf/cvpr/ChenH21}
Xinlei Chen and Kaiming He.
\newblock Exploring simple siamese representation learning.
\newblock In \emph{{IEEE} Conference on Computer Vision and Pattern
  Recognition, {CVPR} 2021, virtual, June 19-25, 2021}, pages 15750--15758.
  Computer Vision Foundation / {IEEE}, 2021.
\newblock URL
  \url{https://openaccess.thecvf.com/content/CVPR2021/html/Chen\_Exploring\_Simple\_Siamese\_Representation\_Learning\_CVPR\_2021\_paper.html}.

\bibitem[Chen et~al.(2020{\natexlab{c}})Chen, Fan, Girshick, and
  He]{DBLP:journals/corr/abs-2003-04297}
Xinlei Chen, Haoqi Fan, Ross~B. Girshick, and Kaiming He.
\newblock Improved baselines with momentum contrastive learning.
\newblock \emph{CoRR}, abs/2003.04297, 2020{\natexlab{c}}.
\newblock URL \url{https://arxiv.org/abs/2003.04297}.

\bibitem[Chen et~al.(2021)Chen, Xie, and He]{DBLP:conf/iccv/ChenXH21}
Xinlei Chen, Saining Xie, and Kaiming He.
\newblock An empirical study of training self-supervised vision transformers.
\newblock In \emph{2021 {IEEE/CVF} International Conference on Computer Vision,
  {ICCV} 2021, Montreal, QC, Canada, October 10-17, 2021}, pages 9620--9629.
  {IEEE}, 2021.
\newblock \doi{10.1109/ICCV48922.2021.00950}.
\newblock URL \url{https://doi.org/10.1109/ICCV48922.2021.00950}.

\bibitem[Cho et~al.(2021)Cho, Mall, Bala, and
  Hariharan]{DBLP:conf/cvpr/ChoMBH21}
Jang~Hyun Cho, Utkarsh Mall, Kavita Bala, and Bharath Hariharan.
\newblock Picie: Unsupervised semantic segmentation using invariance and
  equivariance in clustering.
\newblock In \emph{{IEEE} Conference on Computer Vision and Pattern
  Recognition, {CVPR} 2021, virtual, June 19-25, 2021}, pages 16794--16804.
  Computer Vision Foundation / {IEEE}, 2021.
\newblock URL
  \url{https://openaccess.thecvf.com/content/CVPR2021/html/Cho\_PiCIE\_Unsupervised\_Semantic\_Segmentation\_Using\_Invariance\_and\_Equivariance\_in\_Clustering\_CVPR\_2021\_paper.html}.

\bibitem[Cordts et~al.(2016)Cordts, Omran, Ramos, Rehfeld, Enzweiler, Benenson,
  Franke, Roth, and Schiele]{DBLP:conf/cvpr/CordtsORREBFRS16}
Marius Cordts, Mohamed Omran, Sebastian Ramos, Timo Rehfeld, Markus Enzweiler,
  Rodrigo Benenson, Uwe Franke, Stefan Roth, and Bernt Schiele.
\newblock The cityscapes dataset for semantic urban scene understanding.
\newblock In \emph{2016 {IEEE} Conference on Computer Vision and Pattern
  Recognition, {CVPR} 2016, Las Vegas, NV, USA, June 27-30, 2016}, pages
  3213--3223. {IEEE} Computer Society, 2016.
\newblock \doi{10.1109/CVPR.2016.350}.
\newblock URL \url{https://doi.org/10.1109/CVPR.2016.350}.

\bibitem[Cuturi(2013)]{DBLP:conf/nips/Cuturi13}
Marco Cuturi.
\newblock Sinkhorn distances: Lightspeed computation of optimal transport.
\newblock In Christopher J.~C. Burges, L{\'{e}}on Bottou, Zoubin Ghahramani,
  and Kilian~Q. Weinberger, editors, \emph{Advances in Neural Information
  Processing Systems 26: 27th Annual Conference on Neural Information
  Processing Systems 2013. Proceedings of a meeting held December 5-8, 2013,
  Lake Tahoe, Nevada, United States}, pages 2292--2300, 2013.
\newblock URL
  \url{https://proceedings.neurips.cc/paper/2013/hash/af21d0c97db2e27e13572cbf59eb343d-Abstract.html}.

\bibitem[Doersch et~al.(2015)Doersch, Gupta, and
  Efros]{DBLP:conf/iccv/DoerschGE15}
Carl Doersch, Abhinav Gupta, and Alexei~A. Efros.
\newblock Unsupervised visual representation learning by context prediction.
\newblock In \emph{2015 {IEEE} International Conference on Computer Vision,
  {ICCV} 2015, Santiago, Chile, December 7-13, 2015}, pages 1422--1430. {IEEE}
  Computer Society, 2015.
\newblock \doi{10.1109/ICCV.2015.167}.
\newblock URL \url{https://doi.org/10.1109/ICCV.2015.167}.

\bibitem[Fang et~al.(2019)Fang, Sun, Wang, Gou, Li, and
  Lu]{DBLP:conf/iccv/FangSWGLL19}
Haoshu Fang, Jianhua Sun, Runzhong Wang, Minghao Gou, Yonglu Li, and Cewu Lu.
\newblock Instaboost: Boosting instance segmentation via probability map guided
  copy-pasting.
\newblock In \emph{2019 {IEEE/CVF} International Conference on Computer Vision,
  {ICCV} 2019, Seoul, Korea (South), October 27 - November 2, 2019}, pages
  682--691. {IEEE}, 2019.
\newblock \doi{10.1109/ICCV.2019.00077}.
\newblock URL \url{https://doi.org/10.1109/ICCV.2019.00077}.

\bibitem[Fortunato(2009)]{DBLP:journals/corr/abs-0906-0612}
Santo Fortunato.
\newblock Community detection in graphs.
\newblock \emph{CoRR}, abs/0906.0612, 2009.
\newblock URL \url{http://arxiv.org/abs/0906.0612}.

\bibitem[Gansbeke et~al.(2021)Gansbeke, Vandenhende, Georgoulis, and
  Gool]{DBLP:conf/iccv/GansbekeVGG21}
Wouter~Van Gansbeke, Simon Vandenhende, Stamatios Georgoulis, and Luc~Van Gool.
\newblock Unsupervised semantic segmentation by contrasting object mask
  proposals.
\newblock In \emph{2021 {IEEE/CVF} International Conference on Computer Vision,
  {ICCV} 2021, Montreal, QC, Canada, October 10-17, 2021}, pages 10032--10042.
  {IEEE}, 2021.
\newblock \doi{10.1109/ICCV48922.2021.00990}.
\newblock URL \url{https://doi.org/10.1109/ICCV48922.2021.00990}.

\bibitem[Geiger et~al.(2013)Geiger, Lenz, Stiller, and Urtasun]{Geiger2013IJRR}
Andreas Geiger, Philip Lenz, Christoph Stiller, and Raquel Urtasun.
\newblock Vision meets robotics: The kitti dataset.
\newblock \emph{International Journal of Robotics Research (IJRR)}, 2013.

\bibitem[Geirhos et~al.(2020)Geirhos, Jacobsen, Michaelis, Zemel, Brendel,
  Bethge, and Wichmann]{DBLP:journals/natmi/GeirhosJMZBBW20}
Robert Geirhos, J{\"{o}}rn{-}Henrik Jacobsen, Claudio Michaelis, Richard~S.
  Zemel, Wieland Brendel, Matthias Bethge, and Felix~A. Wichmann.
\newblock Shortcut learning in deep neural networks.
\newblock \emph{Nat. Mach. Intell.}, 2\penalty0 (11):\penalty0 665--673, 2020.
\newblock \doi{10.1038/s42256-020-00257-z}.
\newblock URL \url{https://doi.org/10.1038/s42256-020-00257-z}.

\bibitem[Ghiasi et~al.(2021)Ghiasi, Cui, Srinivas, Qian, Lin, Cubuk, Le, and
  Zoph]{DBLP:conf/cvpr/GhiasiCSQLCLZ21}
Golnaz Ghiasi, Yin Cui, Aravind Srinivas, Rui Qian, Tsung{-}Yi Lin, Ekin~D.
  Cubuk, Quoc~V. Le, and Barret Zoph.
\newblock Simple copy-paste is a strong data augmentation method for instance
  segmentation.
\newblock In \emph{{IEEE} Conference on Computer Vision and Pattern
  Recognition, {CVPR} 2021, virtual, June 19-25, 2021}, pages 2918--2928.
  Computer Vision Foundation / {IEEE}, 2021.
\newblock URL
  \url{https://openaccess.thecvf.com/content/CVPR2021/html/Ghiasi\_Simple\_Copy-Paste\_Is\_a\_Strong\_Data\_Augmentation\_Method\_for\_Instance\_CVPR\_2021\_paper.html}.

\bibitem[Gidaris et~al.(2018)Gidaris, Singh, and
  Komodakis]{DBLP:conf/iclr/GidarisSK18}
Spyros Gidaris, Praveer Singh, and Nikos Komodakis.
\newblock Unsupervised representation learning by predicting image rotations.
\newblock In \emph{6th International Conference on Learning Representations,
  {ICLR} 2018, Vancouver, BC, Canada, April 30 - May 3, 2018, Conference Track
  Proceedings}. OpenReview.net, 2018.
\newblock URL \url{https://openreview.net/forum?id=S1v4N2l0-}.

\bibitem[Godard et~al.(2019)Godard, Aodha, Firman, and
  Brostow]{DBLP:conf/iccv/GodardAFB19}
Cl{\'{e}}ment Godard, Oisin~Mac Aodha, Michael Firman, and Gabriel~J. Brostow.
\newblock Digging into self-supervised monocular depth estimation.
\newblock In \emph{2019 {IEEE/CVF} International Conference on Computer Vision,
  {ICCV} 2019, Seoul, Korea (South), October 27 - November 2, 2019}, pages
  3827--3837. {IEEE}, 2019.
\newblock \doi{10.1109/ICCV.2019.00393}.
\newblock URL \url{https://doi.org/10.1109/ICCV.2019.00393}.

\bibitem[Hadsell et~al.(2006)Hadsell, Chopra, and
  LeCun]{DBLP:conf/cvpr/HadsellCL06}
Raia Hadsell, Sumit Chopra, and Yann LeCun.
\newblock Dimensionality reduction by learning an invariant mapping.
\newblock In \emph{2006 {IEEE} Computer Society Conference on Computer Vision
  and Pattern Recognition {(CVPR} 2006), 17-22 June 2006, New York, NY, {USA}},
  pages 1735--1742. {IEEE} Computer Society, 2006.
\newblock \doi{10.1109/CVPR.2006.100}.
\newblock URL \url{https://doi.org/10.1109/CVPR.2006.100}.

\bibitem[He et~al.(2016)He, Zhang, Ren, and Sun]{He_2016_CVPR}
Kaiming He, Xiangyu Zhang, Shaoqing Ren, and Jian Sun.
\newblock Deep residual learning for image recognition.
\newblock In \emph{Proceedings of the IEEE Conference on Computer Vision and
  Pattern Recognition (CVPR)}, June 2016.

\bibitem[He et~al.(2017)He, Gkioxari, Doll{\'{a}}r, and
  Girshick]{DBLP:conf/iccv/HeGDG17}
Kaiming He, Georgia Gkioxari, Piotr Doll{\'{a}}r, and Ross~B. Girshick.
\newblock Mask {R-CNN}.
\newblock In \emph{{IEEE} International Conference on Computer Vision, {ICCV}
  2017, Venice, Italy, October 22-29, 2017}, pages 2980--2988. {IEEE} Computer
  Society, 2017.
\newblock \doi{10.1109/ICCV.2017.322}.
\newblock URL \url{https://doi.org/10.1109/ICCV.2017.322}.

\bibitem[He et~al.(2020)He, Fan, Wu, Xie, and
  Girshick]{DBLP:conf/cvpr/He0WXG20}
Kaiming He, Haoqi Fan, Yuxin Wu, Saining Xie, and Ross~B. Girshick.
\newblock Momentum contrast for unsupervised visual representation learning.
\newblock In \emph{2020 {IEEE/CVF} Conference on Computer Vision and Pattern
  Recognition, {CVPR} 2020, Seattle, WA, USA, June 13-19, 2020}, pages
  9726--9735. Computer Vision Foundation / {IEEE}, 2020.
\newblock \doi{10.1109/CVPR42600.2020.00975}.
\newblock URL \url{https://doi.org/10.1109/CVPR42600.2020.00975}.

\bibitem[He et~al.(2021)He, Chen, Xie, Li, Doll{\'{a}}r, and
  Girshick]{DBLP:journals/corr/abs-2111-06377}
Kaiming He, Xinlei Chen, Saining Xie, Yanghao Li, Piotr Doll{\'{a}}r, and
  Ross~B. Girshick.
\newblock Masked autoencoders are scalable vision learners.
\newblock \emph{CoRR}, abs/2111.06377, 2021.
\newblock URL \url{https://arxiv.org/abs/2111.06377}.

\bibitem[H{\'{e}}naff et~al.(2021)H{\'{e}}naff, Koppula, Alayrac, van~den Oord,
  Vinyals, and Carreira]{DBLP:conf/iccv/HenaffKAOVC21}
Olivier~J. H{\'{e}}naff, Skanda Koppula, Jean{-}Baptiste Alayrac, A{\"{a}}ron
  van~den Oord, Oriol Vinyals, and Jo{\~{a}}o Carreira.
\newblock Efficient visual pretraining with contrastive detection.
\newblock In \emph{2021 {IEEE/CVF} International Conference on Computer Vision,
  {ICCV} 2021, Montreal, QC, Canada, October 10-17, 2021}, pages 10066--10076.
  {IEEE}, 2021.
\newblock \doi{10.1109/ICCV48922.2021.00993}.
\newblock URL \url{https://doi.org/10.1109/ICCV48922.2021.00993}.

\bibitem[Hoyer et~al.(2021)Hoyer, Dai, Chen, K{\"{o}}ring, Saha, and
  Gool]{DBLP:conf/cvpr/HoyerDCKSG21}
Lukas Hoyer, Dengxin Dai, Yuhua Chen, Adrian K{\"{o}}ring, Suman Saha, and
  Luc~Van Gool.
\newblock Three ways to improve semantic segmentation with self-supervised
  depth estimation.
\newblock In \emph{{IEEE} Conference on Computer Vision and Pattern
  Recognition, {CVPR} 2021, virtual, June 19-25, 2021}, pages 11130--11140.
  Computer Vision Foundation / {IEEE}, 2021.
\newblock URL
  \url{https://openaccess.thecvf.com/content/CVPR2021/html/Hoyer\_Three\_Ways\_To\_Improve\_Semantic\_Segmentation\_With\_Self-Supervised\_Depth\_Estimation\_CVPR\_2021\_paper.html}.

\bibitem[Kayhan and van Gemert(2022)]{DBLP:journals/corr/abs-2205-02887}
Osman~Semih Kayhan and Jan~C. van Gemert.
\newblock Evaluating context for deep object detectors.
\newblock \emph{CoRR}, abs/2205.02887, 2022.
\newblock \doi{10.48550/arXiv.2205.02887}.
\newblock URL \url{https://doi.org/10.48550/arXiv.2205.02887}.

\bibitem[Kirillov et~al.(2019)Kirillov, Girshick, He, and
  Doll{\'{a}}r]{DBLP:conf/cvpr/KirillovGHD19}
Alexander Kirillov, Ross~B. Girshick, Kaiming He, and Piotr Doll{\'{a}}r.
\newblock Panoptic feature pyramid networks.
\newblock In \emph{{IEEE} Conference on Computer Vision and Pattern
  Recognition, {CVPR} 2019, Long Beach, CA, USA, June 16-20, 2019}, pages
  6399--6408. Computer Vision Foundation / {IEEE}, 2019.
\newblock \doi{10.1109/CVPR.2019.00656}.
\newblock URL
  \url{http://openaccess.thecvf.com/content\_CVPR\_2019/html/Kirillov\_Panoptic\_Feature\_Pyramid\_Networks\_CVPR\_2019\_paper.html}.

\bibitem[Lin et~al.(2014)Lin, Maire, Belongie, Hays, Perona, Ramanan,
  Doll{\'{a}}r, and Zitnick]{DBLP:conf/eccv/LinMBHPRDZ14}
Tsung{-}Yi Lin, Michael Maire, Serge~J. Belongie, James Hays, Pietro Perona,
  Deva Ramanan, Piotr Doll{\'{a}}r, and C.~Lawrence Zitnick.
\newblock Microsoft {COCO:} common objects in context.
\newblock In David~J. Fleet, Tom{\'{a}}s Pajdla, Bernt Schiele, and Tinne
  Tuytelaars, editors, \emph{Computer Vision - {ECCV} 2014 - 13th European
  Conference, Zurich, Switzerland, September 6-12, 2014, Proceedings, Part
  {V}}, volume 8693 of \emph{Lecture Notes in Computer Science}, pages
  740--755. Springer, 2014.
\newblock \doi{10.1007/978-3-319-10602-1\_48}.
\newblock URL \url{https://doi.org/10.1007/978-3-319-10602-1\_48}.

\bibitem[Long et~al.(2015)Long, Shelhamer, and Darrell]{7298965}
Jonathan Long, Evan Shelhamer, and Trevor Darrell.
\newblock Fully convolutional networks for semantic segmentation.
\newblock In \emph{2015 IEEE Conference on Computer Vision and Pattern
  Recognition (CVPR)}, pages 3431--3440, 2015.
\newblock \doi{10.1109/CVPR.2015.7298965}.

\bibitem[Noroozi and Favaro(2016)]{DBLP:conf/eccv/NorooziF16}
Mehdi Noroozi and Paolo Favaro.
\newblock Unsupervised learning of visual representations by solving jigsaw
  puzzles.
\newblock In Bastian Leibe, Jiri Matas, Nicu Sebe, and Max Welling, editors,
  \emph{Computer Vision - {ECCV} 2016 - 14th European Conference, Amsterdam,
  The Netherlands, October 11-14, 2016, Proceedings, Part {VI}}, volume 9910 of
  \emph{Lecture Notes in Computer Science}, pages 69--84. Springer, 2016.
\newblock \doi{10.1007/978-3-319-46466-4\_5}.
\newblock URL \url{https://doi.org/10.1007/978-3-319-46466-4\_5}.

\bibitem[Pathak et~al.(2017)Pathak, Girshick, Doll{\'{a}}r, Darrell, and
  Hariharan]{DBLP:conf/cvpr/PathakGDDH17}
Deepak Pathak, Ross~B. Girshick, Piotr Doll{\'{a}}r, Trevor Darrell, and
  Bharath Hariharan.
\newblock Learning features by watching objects move.
\newblock In \emph{2017 {IEEE} Conference on Computer Vision and Pattern
  Recognition, {CVPR} 2017, Honolulu, HI, USA, July 21-26, 2017}, pages
  6024--6033. {IEEE} Computer Society, 2017.
\newblock \doi{10.1109/CVPR.2017.638}.
\newblock URL \url{https://doi.org/10.1109/CVPR.2017.638}.

\bibitem[Pinheiro et~al.(2020)Pinheiro, Almahairi, Benmalek, Golemo, and
  Courville]{DBLP:conf/nips/PinheiroABGC20}
Pedro~O. Pinheiro, Amjad Almahairi, Ryan~Y. Benmalek, Florian Golemo, and
  Aaron~C. Courville.
\newblock Unsupervised learning of dense visual representations.
\newblock In Hugo Larochelle, Marc'Aurelio Ranzato, Raia Hadsell,
  Maria{-}Florina Balcan, and Hsuan{-}Tien Lin, editors, \emph{Advances in
  Neural Information Processing Systems 33: Annual Conference on Neural
  Information Processing Systems 2020, NeurIPS 2020, December 6-12, 2020,
  virtual}, 2020.
\newblock URL
  \url{https://proceedings.neurips.cc/paper/2020/hash/3000311ca56a1cb93397bc676c0b7fff-Abstract.html}.

\bibitem[Ren et~al.(2015)Ren, He, Girshick, and Sun]{DBLP:conf/nips/RenHGS15}
Shaoqing Ren, Kaiming He, Ross~B. Girshick, and Jian Sun.
\newblock Faster {R-CNN:} towards real-time object detection with region
  proposal networks.
\newblock In Corinna Cortes, Neil~D. Lawrence, Daniel~D. Lee, Masashi Sugiyama,
  and Roman Garnett, editors, \emph{Advances in Neural Information Processing
  Systems 28: Annual Conference on Neural Information Processing Systems 2015,
  December 7-12, 2015, Montreal, Quebec, Canada}, pages 91--99, 2015.
\newblock URL
  \url{https://proceedings.neurips.cc/paper/2015/hash/14bfa6bb14875e45bba028a21ed38046-Abstract.html}.

\bibitem[Rosvall et~al.(2009)Rosvall, Axelsson, and Bergstrom]{Rosvall_2009}
M.~Rosvall, D.~Axelsson, and C.~T. Bergstrom.
\newblock The map equation.
\newblock \emph{The European Physical Journal Special Topics}, 178\penalty0
  (1):\penalty0 13--23, nov 2009.
\newblock \doi{10.1140/epjst/e2010-01179-1}.
\newblock URL \url{https://doi.org/10.1140%2Fepjst%2Fe2010-01179-1}.

\bibitem[Russakovsky et~al.(2015)Russakovsky, Deng, Su, Krause, Satheesh, Ma,
  Huang, Karpathy, Khosla, Bernstein, Berg, and Fei-Fei]{ILSVRC15}
Olga Russakovsky, Jia Deng, Hao Su, Jonathan Krause, Sanjeev Satheesh, Sean Ma,
  Zhiheng Huang, Andrej Karpathy, Aditya Khosla, Michael Bernstein,
  Alexander~C. Berg, and Li~Fei-Fei.
\newblock {ImageNet Large Scale Visual Recognition Challenge}.
\newblock \emph{International Journal of Computer Vision (IJCV)}, 115\penalty0
  (3):\penalty0 211--252, 2015.
\newblock \doi{10.1007/s11263-015-0816-y}.

\bibitem[Selvaraju et~al.(2021)Selvaraju, Desai, Johnson, and
  Naik]{DBLP:conf/cvpr/SelvarajuD0N21}
Ramprasaath~R. Selvaraju, Karan Desai, Justin Johnson, and Nikhil Naik.
\newblock Casting your model: Learning to localize improves self-supervised
  representations.
\newblock In \emph{{IEEE} Conference on Computer Vision and Pattern
  Recognition, {CVPR} 2021, virtual, June 19-25, 2021}, pages 11058--11067.
  Computer Vision Foundation / {IEEE}, 2021.
\newblock URL
  \url{https://openaccess.thecvf.com/content/CVPR2021/html/Selvaraju\_CASTing\_Your\_Model\_Learning\_To\_Localize\_Improves\_Self-Supervised\_Representations\_CVPR\_2021\_paper.html}.

\bibitem[Silberman et~al.(2012)Silberman, Hoiem, Kohli, and
  Fergus]{DBLP:conf/eccv/SilbermanHKF12}
Nathan Silberman, Derek Hoiem, Pushmeet Kohli, and Rob Fergus.
\newblock Indoor segmentation and support inference from {RGBD} images.
\newblock In Andrew~W. Fitzgibbon, Svetlana Lazebnik, Pietro Perona, Yoichi
  Sato, and Cordelia Schmid, editors, \emph{Computer Vision - {ECCV} 2012 -
  12th European Conference on Computer Vision, Florence, Italy, October 7-13,
  2012, Proceedings, Part {V}}, volume 7576 of \emph{Lecture Notes in Computer
  Science}, pages 746--760. Springer, 2012.
\newblock \doi{10.1007/978-3-642-33715-4\_54}.
\newblock URL \url{https://doi.org/10.1007/978-3-642-33715-4\_54}.

\bibitem[Tian et~al.(2020)Tian, Sun, Poole, Krishnan, Schmid, and
  Isola]{DBLP:conf/nips/Tian0PKSI20}
Yonglong Tian, Chen Sun, Ben Poole, Dilip Krishnan, Cordelia Schmid, and
  Phillip Isola.
\newblock What makes for good views for contrastive learning?
\newblock In Hugo Larochelle, Marc'Aurelio Ranzato, Raia Hadsell,
  Maria{-}Florina Balcan, and Hsuan{-}Tien Lin, editors, \emph{Advances in
  Neural Information Processing Systems 33: Annual Conference on Neural
  Information Processing Systems 2020, NeurIPS 2020, December 6-12, 2020,
  virtual}, 2020.
\newblock URL
  \url{https://proceedings.neurips.cc/paper/2020/hash/4c2e5eaae9152079b9e95845750bb9ab-Abstract.html}.

\bibitem[Vondrick et~al.(2018)Vondrick, Shrivastava, Fathi, Guadarrama, and
  Murphy]{DBLP:conf/eccv/VondrickSFGM18}
Carl Vondrick, Abhinav Shrivastava, Alireza Fathi, Sergio Guadarrama, and Kevin
  Murphy.
\newblock Tracking emerges by colorizing videos.
\newblock In Vittorio Ferrari, Martial Hebert, Cristian Sminchisescu, and Yair
  Weiss, editors, \emph{Computer Vision - {ECCV} 2018 - 15th European
  Conference, Munich, Germany, September 8-14, 2018, Proceedings, Part {XIII}},
  volume 11217 of \emph{Lecture Notes in Computer Science}, pages 402--419.
  Springer, 2018.
\newblock \doi{10.1007/978-3-030-01261-8\_24}.
\newblock URL \url{https://doi.org/10.1007/978-3-030-01261-8\_24}.

\bibitem[Wang et~al.(2022)Wang, Wang, Wei, Yuille, and
  Shen]{DBLP:journals/corr/abs-2203-11709}
Feng Wang, Huiyu Wang, Chen Wei, Alan~L. Yuille, and Wei Shen.
\newblock {CP2:} copy-paste contrastive pretraining for semantic segmentation.
\newblock \emph{CoRR}, abs/2203.11709, 2022.
\newblock \doi{10.48550/arXiv.2203.11709}.
\newblock URL \url{https://doi.org/10.48550/arXiv.2203.11709}.

\bibitem[Wang and Gupta(2015)]{DBLP:conf/iccv/WangG15}
Xiaolong Wang and Abhinav Gupta.
\newblock Unsupervised learning of visual representations using videos.
\newblock In \emph{2015 {IEEE} International Conference on Computer Vision,
  {ICCV} 2015, Santiago, Chile, December 7-13, 2015}, pages 2794--2802. {IEEE}
  Computer Society, 2015.
\newblock \doi{10.1109/ICCV.2015.320}.
\newblock URL \url{https://doi.org/10.1109/ICCV.2015.320}.

\bibitem[Wang et~al.(2021)Wang, Zhang, Shen, Kong, and
  Li]{DBLP:conf/cvpr/WangZSKL21}
Xinlong Wang, Rufeng Zhang, Chunhua Shen, Tao Kong, and Lei Li.
\newblock Dense contrastive learning for self-supervised visual pre-training.
\newblock In \emph{{IEEE} Conference on Computer Vision and Pattern
  Recognition, {CVPR} 2021, virtual, June 19-25, 2021}, pages 3024--3033.
  Computer Vision Foundation / {IEEE}, 2021.
\newblock URL
  \url{https://openaccess.thecvf.com/content/CVPR2021/html/Wang\_Dense\_Contrastive\_Learning\_for\_Self-Supervised\_Visual\_Pre-Training\_CVPR\_2021\_paper.html}.

\bibitem[Wu et~al.(2019)Wu, Kirillov, Massa, Lo, and
  Girshick]{wu2019detectron2}
Yuxin Wu, Alexander Kirillov, Francisco Massa, Wan-Yen Lo, and Ross Girshick.
\newblock Detectron2.
\newblock \url{https://github.com/facebookresearch/detectron2}, 2019.

\bibitem[Xie et~al.(2021{\natexlab{a}})Xie, Zhan, Liu, Ong, and
  Loy]{xie2021unsupervised}
Jiahao Xie, Xiaohang Zhan, Ziwei Liu, Yew~Soon Ong, and Chen~Change Loy.
\newblock Unsupervised object-level representation learning from scene images.
\newblock In \emph{NeurIPS}, 2021{\natexlab{a}}.

\bibitem[Xie et~al.(2021{\natexlab{b}})Xie, Lin, Zhang, Cao, Lin, and
  Hu]{DBLP:conf/cvpr/XieL00L021}
Zhenda Xie, Yutong Lin, Zheng Zhang, Yue Cao, Stephen Lin, and Han Hu.
\newblock Propagate yourself: Exploring pixel-level consistency for
  unsupervised visual representation learning.
\newblock In \emph{{IEEE} Conference on Computer Vision and Pattern
  Recognition, {CVPR} 2021, virtual, June 19-25, 2021}, pages 16684--16693.
  Computer Vision Foundation / {IEEE}, 2021{\natexlab{b}}.
\newblock URL
  \url{https://openaccess.thecvf.com/content/CVPR2021/html/Xie\_Propagate\_Yourself\_Exploring\_Pixel-Level\_Consistency\_for\_Unsupervised\_Visual\_Representation\_Learning\_CVPR\_2021\_paper.html}.

\bibitem[Xu et~al.(2021)Xu, Zhang, Zhang, and
  Tao]{DBLP:journals/corr/abs-2111-12309}
Yufei Xu, Qiming Zhang, Jing Zhang, and Dacheng Tao.
\newblock Regioncl: Can simple region swapping contribute to contrastive
  learning?
\newblock \emph{CoRR}, abs/2111.12309, 2021.
\newblock URL \url{https://arxiv.org/abs/2111.12309}.

\bibitem[Yun et~al.(2019)Yun, Han, Chun, Oh, Yoo, and
  Choe]{DBLP:conf/iccv/YunHCOYC19}
Sangdoo Yun, Dongyoon Han, Sanghyuk Chun, Seong~Joon Oh, Youngjoon Yoo, and
  Junsuk Choe.
\newblock Cutmix: Regularization strategy to train strong classifiers with
  localizable features.
\newblock In \emph{2019 {IEEE/CVF} International Conference on Computer Vision,
  {ICCV} 2019, Seoul, Korea (South), October 27 - November 2, 2019}, pages
  6022--6031. {IEEE}, 2019.
\newblock \doi{10.1109/ICCV.2019.00612}.
\newblock URL \url{https://doi.org/10.1109/ICCV.2019.00612}.

\bibitem[Zbontar et~al.(2021)Zbontar, Jing, Misra, LeCun, and
  Deny]{DBLP:conf/icml/ZbontarJMLD21}
Jure Zbontar, Li~Jing, Ishan Misra, Yann LeCun, and St{\'{e}}phane Deny.
\newblock Barlow twins: Self-supervised learning via redundancy reduction.
\newblock In Marina Meila and Tong Zhang, editors, \emph{Proceedings of the
  38th International Conference on Machine Learning, {ICML} 2021, 18-24 July
  2021, Virtual Event}, volume 139 of \emph{Proceedings of Machine Learning
  Research}, pages 12310--12320. {PMLR}, 2021.
\newblock URL \url{http://proceedings.mlr.press/v139/zbontar21a.html}.

\bibitem[Zhang et~al.(2021)Zhang, Torr, Ranftl, and Richter]{zhang2021looking}
Feihu Zhang, Philip Torr, Rene Ranftl, and Stephan Richter.
\newblock Looking beyond single images for contrastive semantic segmentation
  learning.
\newblock \emph{Advances in Neural Information Processing Systems}, 34, 2021.

\bibitem[Zhang et~al.(2020)Zhang, Tseng, and Kreiman]{DBLP:conf/cvpr/ZhangTK20}
Mengmi Zhang, Claire Tseng, and Gabriel Kreiman.
\newblock Putting visual object recognition in context.
\newblock In \emph{2020 {IEEE/CVF} Conference on Computer Vision and Pattern
  Recognition, {CVPR} 2020, Seattle, WA, USA, June 13-19, 2020}, pages
  12982--12991. Computer Vision Foundation / {IEEE}, 2020.
\newblock \doi{10.1109/CVPR42600.2020.01300}.
\newblock URL
  \url{https://openaccess.thecvf.com/content\_CVPR\_2020/html/Zhang\_Putting\_Visual\_Object\_Recognition\_in\_Context\_CVPR\_2020\_paper.html}.

\bibitem[Zhang et~al.(2016)Zhang, Isola, and Efros]{DBLP:conf/eccv/ZhangIE16}
Richard Zhang, Phillip Isola, and Alexei~A. Efros.
\newblock Colorful image colorization.
\newblock In Bastian Leibe, Jiri Matas, Nicu Sebe, and Max Welling, editors,
  \emph{Computer Vision - {ECCV} 2016 - 14th European Conference, Amsterdam,
  The Netherlands, October 11-14, 2016, Proceedings, Part {III}}, volume 9907
  of \emph{Lecture Notes in Computer Science}, pages 649--666. Springer, 2016.
\newblock \doi{10.1007/978-3-319-46487-9\_40}.
\newblock URL \url{https://doi.org/10.1007/978-3-319-46487-9\_40}.

\bibitem[Zhang and Maire(2020)]{DBLP:conf/nips/ZhangM20}
Xiao Zhang and Michael Maire.
\newblock Self-supervised visual representation learning from hierarchical
  grouping.
\newblock In Hugo Larochelle, Marc'Aurelio Ranzato, Raia Hadsell,
  Maria{-}Florina Balcan, and Hsuan{-}Tien Lin, editors, \emph{Advances in
  Neural Information Processing Systems 33: Annual Conference on Neural
  Information Processing Systems 2020, NeurIPS 2020, December 6-12, 2020,
  virtual}, 2020.
\newblock URL
  \url{https://proceedings.neurips.cc/paper/2020/hash/c1502ae5a4d514baec129f72948c266e-Abstract.html}.

\bibitem[Zhao et~al.(2020)Zhao, Sun, Zhang, Tang, and
  Qian]{DBLP:journals/corr/abs-2003-06620}
Chaoqiang Zhao, Qiyu Sun, Chongzhen Zhang, Yang Tang, and Feng Qian.
\newblock Monocular depth estimation based on deep learning: An overview.
\newblock \emph{CoRR}, abs/2003.06620, 2020.
\newblock URL \url{https://arxiv.org/abs/2003.06620}.

\bibitem[Zhao et~al.(2021)Zhao, Wu, Lau, and Lin]{DBLP:conf/aaai/ZhaoWLL21}
Nanxuan Zhao, Zhirong Wu, Rynson W.~H. Lau, and Stephen Lin.
\newblock Distilling localization for self-supervised representation learning.
\newblock In \emph{Thirty-Fifth {AAAI} Conference on Artificial Intelligence,
  {AAAI} 2021, Thirty-Third Conference on Innovative Applications of Artificial
  Intelligence, {IAAI} 2021, The Eleventh Symposium on Educational Advances in
  Artificial Intelligence, {EAAI} 2021, Virtual Event, February 2-9, 2021},
  pages 10990--10998. {AAAI} Press, 2021.
\newblock URL \url{https://ojs.aaai.org/index.php/AAAI/article/view/17312}.

\bibitem[Zhou et~al.(2017)Zhou, Brown, Snavely, and
  Lowe]{DBLP:conf/cvpr/ZhouBSL17}
Tinghui Zhou, Matthew Brown, Noah Snavely, and David~G. Lowe.
\newblock Unsupervised learning of depth and ego-motion from video.
\newblock In \emph{2017 {IEEE} Conference on Computer Vision and Pattern
  Recognition, {CVPR} 2017, Honolulu, HI, USA, July 21-26, 2017}, pages
  6612--6619. {IEEE} Computer Society, 2017.
\newblock \doi{10.1109/CVPR.2017.700}.
\newblock URL \url{https://doi.org/10.1109/CVPR.2017.700}.

\end{thebibliography}

\clearpage
\setcounter{section}{0}
\renewcommand{\thesubsection}{\Alph{subsection}.}
\section*{Supplementary Materials}
\subsection{Iterative InfoMap for Community Detection}
\begin{figure}[H]
\centering
\begin{tabular}{cc}
\centering
\bmvaHangBox{\includegraphics[width=5cm]{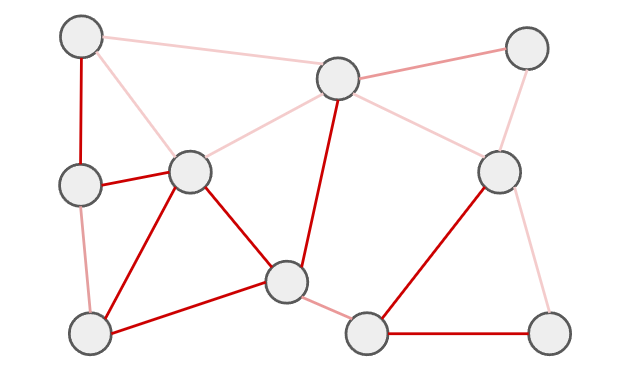}}&
\bmvaHangBox{\includegraphics[width=5cm]{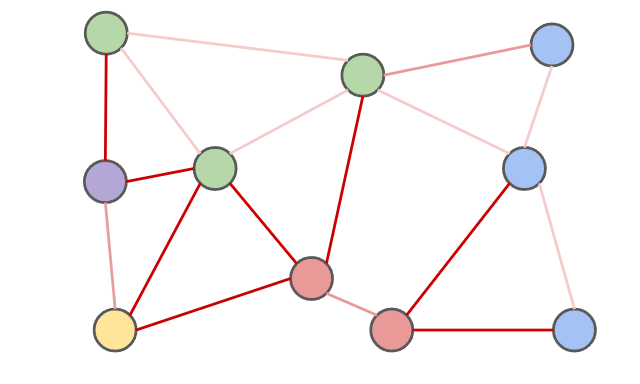}}\vspace{-0.2cm}\\
(a) &(b)\\
\bmvaHangBox{\includegraphics[width=5cm]{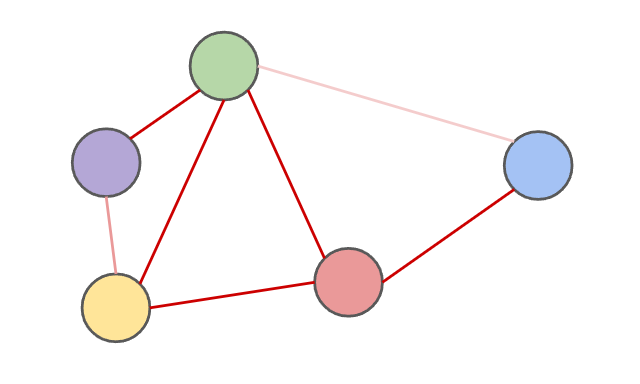}}&
\bmvaHangBox{\includegraphics[width=5cm]{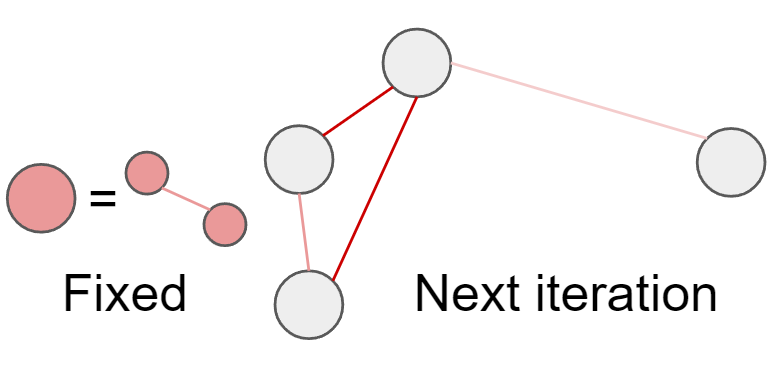}}\\
(c) &(d)\\
\end{tabular}
\caption{(a) Weighted boundary graph with weakly connected edges indicated in deep red. (b) Community detection result by InfoMap~\cite{Rosvall_2009}. The desired community number is half of the node number (c) Community graph (d) Communities with all edges larger than a threshold are fixed and removed. The ultimate community ids are determined for the initial nodes inside the fixed communities. The rest of the graph is fed to the next iteration.}
\label{cd}
\end{figure}

To group coherent region, we first build the region adjacency graph from superpixels. The edges are weighted by the region connectivity measured by Eqn. ~\ref{do} and Eqn.~\ref{ds}. To group strongly connected nodes while separate them from weakly connected nodes from other group, we employ community detection algorithm~\cite{DBLP:journals/corr/abs-0906-0612,DBLP:conf/iwcia/BrowetAD11}. Based on our experiment, we find that directly executing InfoMap~\cite{Rosvall_2009} once with our spatial boundary graph fails to detect multi-scale communities. Therefore, we apply InfoMap progressively. In each iteration, the target number of communities is set to half of the current nodes in the current graph. A new graph is built by treating the detected communities as nodes and its edge weights are computed by the average weight of all original edges linking the two communities. The communities with all outward edges larger than $T_e$ are fixed and removed from the new graph. The initial nodes inside them are assigned with the ultimate community id. The rest of the graph is then fed to the next iteration until only one community is left. The procedure is illustrated in Figure\ref{cd} and the pseudo code is given in Algorithm~\ref{cda}.

\renewcommand{\algorithmicrequire}{\textbf{Input:}}
\renewcommand{\algorithmicensure}{\textbf{Output:}} 
\begin{algorithm}[H]
	\caption{Community Detection} 
	\label{cda} 
	\begin{algorithmic}
		\REQUIRE graph $initialG$, threshold $T_e$, function $InfoMap(graph,\ desired\ community\ num)$
		\ENSURE graph $initialG$ with ultimate community id for its nodes
		
		\FORALL{$n \in initialG.nodes$}
		\STATE \textcolor[RGB]{0,200,0}{\# save the initial nodes contained in the current node}
		\STATE $n.initial\_nodes \gets [n]$ 
		\ENDFOR
		\STATE \textcolor[RGB]{0,200,0}{\# next ultimate community id to assign}
		\STATE $next\_id = 0$
		\STATE $G = initialG$
		\WHILE{$len(G.nodes) > 1$} 
		\STATE \textcolor[RGB]{0,200,0}{\# run InfoMap once to get roughly M//2 temporary communities}
		\STATE $community = InfoMap(G,\ N//2)$
		\FORALL{$n \in G.nodes$}
		\STATE \textcolor[RGB]{0,200,0}{\# assign temporary community id}
		\STATE $n.c \gets community[n]$
		\ENDFOR
		\STATE \textcolor[RGB]{0,200,0}{\# build new graph by treating temporary communities as nodes}
		\STATE $nextG \gets Empty\ Graph$
		\FORALL {$c \in unique(community)$}
		\STATE $nextG.insert\_node(c)$
		\STATE $c.initial\_nodes \gets all\ n.initial\_nodes\ where\ n \in G.nodes\ and\ n.c = c$
		\ENDFOR
		\STATE \textcolor[RGB]{0,200,0}{\# add new edges by averaging original edges between each pair of communities}
		\FORALL {$a \in nextG.nodes\ and\ b \in nextG.nodes$}
		\STATE $weight \gets avg(edge_{nm}.weight)\ where\ edge_{nm} \in G.edges\ and\ n.c = a\ and\ m.c = b$
		\IF{$weight>0$}
		\STATE $edge_{ab}.weight\gets weight$
		\STATE $nextG.insert\_edge(edge_{ab})$
		\ENDIF
		\ENDFOR
		\STATE \textcolor[RGB]{0,200,0}{\# remove "separated community" and assign ultimate community for initial nodes in it}
		\FORALL{$a \in newG$}
		\IF{$all\ edge_{ab}>T_e\ where\ b \in nextG.nodes\ and\ edge_{ab} \in nextG.edges$ }
		\FORALL{$n \in a.initial\_nodes$}
		\STATE $n.ultimate\_community \gets next\_id$
		\ENDFOR
		\STATE $next\_id++$
		\STATE $nextG.remove(a)$
		
		\ENDIF
		\ENDFOR
		\STATE $G \gets nextG$
		\ENDWHILE 
		\RETURN $initialG$
	\end{algorithmic} 
\end{algorithm}

\subsection{Quantitative Evaluation for Pixel Grouping}
To quantify the quality of our coherent pixels grouping result, we split the instance label in Cityscapes~\cite{DBLP:conf/cvpr/CordtsORREBFRS16} $train$ into only connected regions as the ground truth masks. We compute the mIoU in two ways in comparison with owt-ucm~\cite{DBLP:journals/pami/ArbelaezMFM11} and the performance is reported in Table \ref{region quality}. When using each GT mask as the query to fetch the closest mask among the regions, our pixel grouping method produces higher mIoU. When using the optimal bilateral match of the grouping result and GT, our method falls behind because of generating many tiny pieces. 

\begin{table}[H]
  \centering
  \begin{tabular}{@{}lcc@{}}
    \toprule
    Method & mIoU(GT query) & mIoU(bilateral match)\\
    \midrule
    Ours & 10.80 & 3.86 \\
    otw-ucm(0.1)~\cite{DBLP:journals/pami/ArbelaezMFM11} & 4.90 & 4.89\\
    otw-ucm(0.05)~\cite{DBLP:journals/pami/ArbelaezMFM11} & 8.52& 8.08 \\
    otw-ucm(0.01)~\cite{DBLP:journals/pami/ArbelaezMFM11} & 7.94& 5.30 \\
    \bottomrule
  \end{tabular}
  \caption{Pixel grouping mIoU by GT query or bilateral match. For owt-ucm~\cite{DBLP:journals/pami/ArbelaezMFM11}, boundaries under the strength threshold will be filtered.}
  \label{region quality}
\end{table}

\subsection{Training Image Synthesis by Copy-Paste}
The pseudo code is given in Algorithm~\ref{cpa} to illustrate the procedure of our training image synthesis by copy-paste.

\begin{algorithm}[H]
	\caption{Copy-Paste} 
	\label{cpa} 
	\begin{algorithmic}
		\REQUIRE num of sample image $M$, size threshold $T$, height threshold $h_t$, expectation $[e_1,e_2]$
		\ENSURE Synthesized images with cross-context correspondences
		\STATE $synthesized\_images \gets []$
		\STATE $I\gets sampling\ M\ images\ with\ depth\ and\ region\ proposals\ from\ dataset$
		\STATE $R\gets all\ depth\ coherent\ regions \in I\ above\ size\ T$
		\STATE \textcolor[RGB]{0,200,0}{\# two rounds for image synthesis with $e_1 < e_2$}
		\STATE \textcolor[RGB]{0,200,0}{\# $e$ indicates the trade-off between showing more background or external regions}
		\FOR{$iter\ in\ range(2)$}
		\FORALL{$img \in I$}
		\STATE $img = deepcopy(img)$
		\STATE $R_{cp} \gets sampling\ e[iter]*len(R)/M\ regions\ from\ R$
		\FORALL{$r \in R_{cp}$}
		\STATE \textcolor[RGB]{0,200,0}{\# resize, color jitter, flip, the position of r will be adjusted accordingly}
		\STATE $r, r.position\_height \gets Augmentation(r, r.position\_height)$
		\STATE $x \gets random(0,img.width)$
		\STATE $y \gets random(r.position\_height-h_t,r.position\_height+h_t)$
		\STATE \textcolor[RGB]{0,200,0}{\# DepthMix copy-paste}
		\STATE $copy\ paste\ mask\odot r\ on\ img\ at\ (x,y)\ where\ mask = (r_{depth}<img_{depth})$
		\ENDFOR
		\STATE $synthesized\_images.append(img)$
		\ENDFOR
		\ENDFOR
		\RETURN $synthesized\_images$

	\end{algorithmic} 
\end{algorithm}

\subsection{Implementation Details}
\paragraph{Hyperparameters}For depth estimation, we used the default setting in Monodepth2~\cite{DBLP:conf/iccv/GodardAFB19}. For 3D adjacency grouping, the number of superpixels was 10000 and  $w_{ocln}$, $w_{sup}$, $b$ in Eqn. ~\ref{do}, ~\ref{ds} and ~\ref{W12} for computing the edge weight are $48.0$, $200.0$, $-4.0$, respectively. We used the default parameters of infomap~\cite{Rosvall_2009}. We used all images in a batch for training image synthesis by copy-paste. The expectation $e$ for copy-paste is set to $1$ in the first round and $2$ in the second round. The region size threshold was 16 for height and 6 for width. Copy-paste position height threshold $h_t$ was 16.

\paragraph{Training}We trained the model for 40 epochs on a single 16GB V100. 8 source images were randomly sampled per iteration, resulting in 16 augmented images as a batch. To speed up training we generated synthesized data equivalent to 2 epochs beforehand and trained on them repeatedly. During training, resized crop, horizontal flip, and color augmentation are applied to full images. The model was updated by the SGD optimizer with momentum $0.9$ and weight decay $1e^{-4}$. The initial learning rate was warmed up to $1e^{-1}$ in the first 2 epochs and then decayed to $1e^{-5}$ via cosine annealing. We held $1000$ prototypes and set temperature $\tau$ to $0.1$ in SwAV~\cite{DBLP:conf/nips/CaronMMGBJ20}. We sampled around 288k features for contrasting per iteration and the budget was allocated to pixel-level and region-level samples equally if $1>\lambda>0$ where $\lambda$ is the loss weight that balance the loss from the two types of positive samples in Eqn. ~\ref{L}, i.e. $\mathcal{L} = \mathcal{L}_{pixel} + (1-\lambda) \mathcal{L}_{region}$. 

\subsection{Evaluation Setting}
\subsubsection*{Unsupervised semantic segmentation}
Few works try to address the challenging urban-scene semantic segmentation in unsupervised learning. We compare our results to the recently introduced PiCIE~\cite{DBLP:conf/cvpr/ChoMBH21}, which achieved the previous state-of-art in this scenario. Noted that, in the original paper, PiCIE is trained with ResNet-18 initialized by ImageNet pre-trained model and their results are evaluated on raw 27 classes instead of the conventional 19 evaluated classes on Cityscapes~\cite{DBLP:conf/cvpr/CordtsORREBFRS16}, we retrained PiCIE with the equivalent setting to ours, e.g. architecture, batch size using the official code. The final performance is computed based on the optimal match between the predicted masks and the ground truth.

\subsubsection*{Semi-supervised segmentation}
We fine-tuned the pre-trained models on Cityscapes~\cite{DBLP:conf/cvpr/CordtsORREBFRS16} and KITTI~\cite{Geiger2013IJRR} for semantic and instance segmentation using semantic FPN~\cite{DBLP:conf/cvpr/KirillovGHD19} and Mask R-CNN~\cite{DBLP:conf/iccv/HeGDG17} implemented by detectron2~\cite{wu2019detectron2}. The batch size was 16 and the fine-tuning on Cityscapes and KITTI took 30k and 6k iterations, respectively. We set a smaller initial learning rate $1e^{-2}$ for all pre-trained parameters. For other randomly initialized parameters, the learning rate was tuned from $3e^{-2}$ to $1e^{-1}$. We smoothly fused these parameters by a cosine decay scheduler with a final learning rate of $1e^{-5}$. This strategy worked well for all pre-trained models. We used random scale, random crop, and color augmentation. We repeated the training for 3 times and report the average performance.

\subsection{Training SwAV on Cityscapes}

\begin{table}[H]
\centering
\begin{tabular}{@{}llc@{}}
\toprule
Method & Training data& mIoU \\
\midrule
scratch & - & 37.85 \\
SwAV~\cite{DBLP:conf/nips/CaronMMGBJ20}& CS-raw & 31.92\\
SwAV~\cite{DBLP:conf/nips/CaronMMGBJ20}& CS-patch & 37.88\\
SwAV~\cite{DBLP:conf/nips/CaronMMGBJ20}& CS-object & 39.63\\
Ours($\lambda=0$)&CS-raw& 46.91\\
Ours($\lambda=0.5$)&CS-raw& 48.87\\
Ours($\lambda=1$)&CS-raw& 48.55\\
supervised & CS-object & 39.28\\
supervised & ImageNet~\cite{ILSVRC15} & 48.33\\
\bottomrule
\end{tabular}
\caption{Effect of different pre-processing of Cityscapes~\cite{DBLP:conf/cvpr/CordtsORREBFRS16} for SwAV~\cite{DBLP:conf/nips/CaronMMGBJ20}. Measurement is based on mIoU by fine-tuning on $1/16$ $train$ set of Cityscapes for semantic segmentation. CS-raw: full-view Cityscapes images; CS-patch: patchified Cityscapes images; CS-object: object-centric Cityscapes images}
\label{swav}
\end{table}

Because the original SwAV~\cite{DBLP:conf/nips/CaronMMGBJ20} is designed on object-centric ImageNet, we did an experiment to study its behaviors on Cityscapes~\cite{DBLP:conf/cvpr/CordtsORREBFRS16} and investigate the best practice to train SwAV on urban scene data in the same setting as ours for comparison. With the 2975 images in Cityscapes~\cite{DBLP:conf/cvpr/CordtsORREBFRS16} $train$ set, we use the raw $384\times 768$ image, or $128\times 128$ patch images by splitting a raw image into 18 equal squares, or object-centric images generated based on ground truth label to train SwAV. To the maximum on our 16GB V100 GPU, the batch size is 16 for raw images and 288 for patch and object-centric images. From Table \ref{swav}, We see that object-centric prior is critical for SwAV. Splitting the images into patches is an approximation to the object-centric images without labels. Using the raw images of complex scenes leads to deteriorated representations . Overall, the gaining by SwAV may be trivial while our method surpasses ImageNet~\cite{ILSVRC15} pre-training, showing that copy-pasting coherent depth regions introduces effective constraints and yields better representations.

\subsection{More Visualizations}

\begin{figure}[H]
\centering
\begin{tabular}{ccccccc}
\centering
\bmvaHangBox{\includegraphics[width=1.4cm]{fig/pca/frankfurt_000000_000576_leftImg8bit.png}}&
\bmvaHangBox{\includegraphics[width=1.4cm]{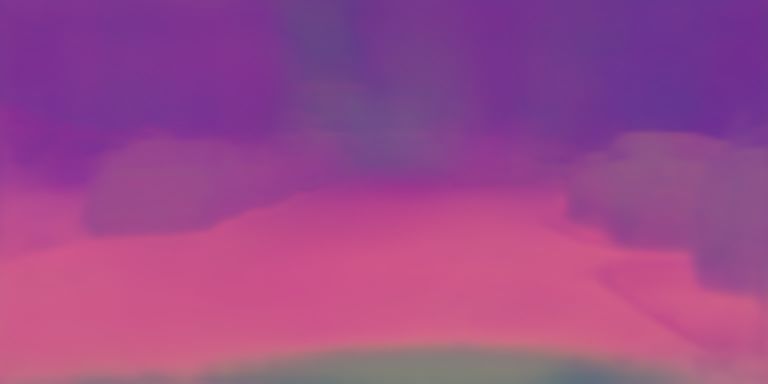}}&
\bmvaHangBox{\includegraphics[width=1.4cm]{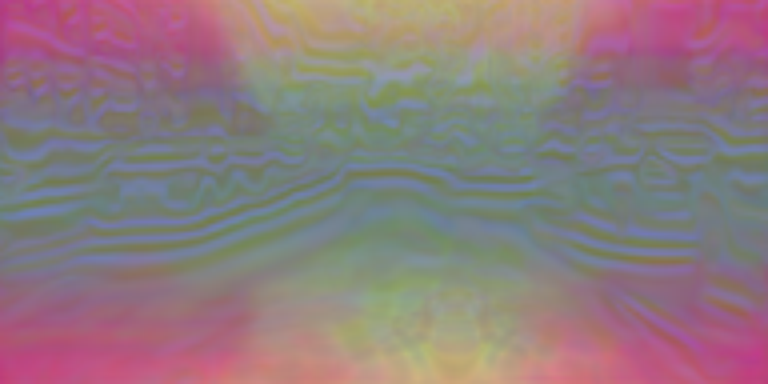}}&
\bmvaHangBox{\includegraphics[width=1.4cm]{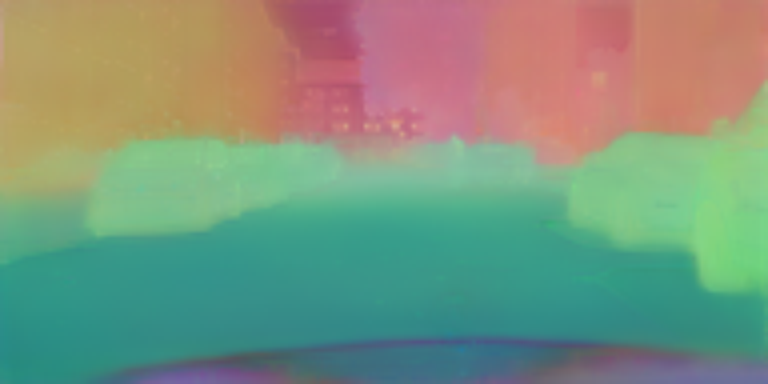}}&
\bmvaHangBox{\includegraphics[width=1.4cm]{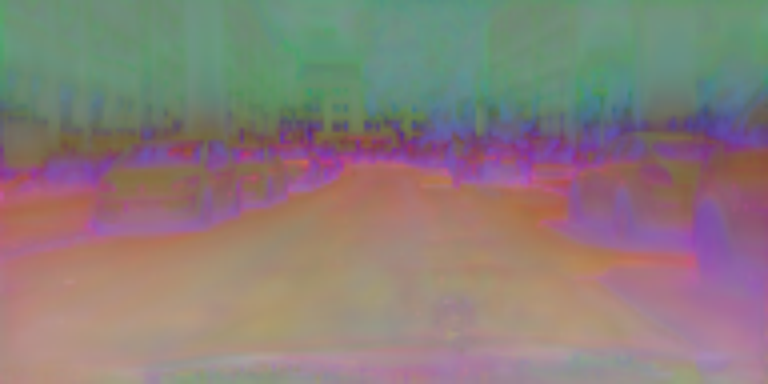}}&
\bmvaHangBox{\includegraphics[width=1.4cm]{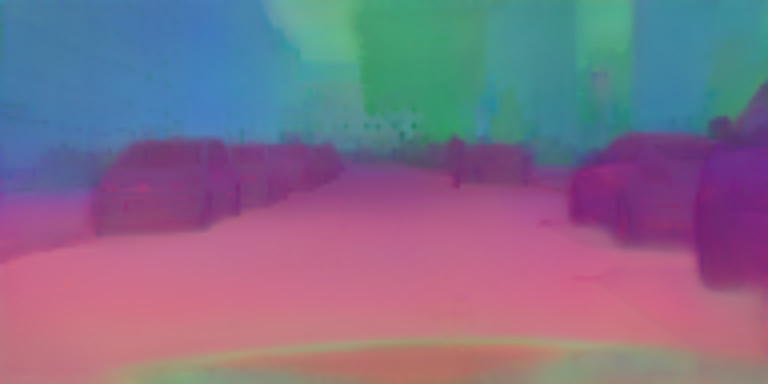}}&
\bmvaHangBox{\includegraphics[width=1.4cm]{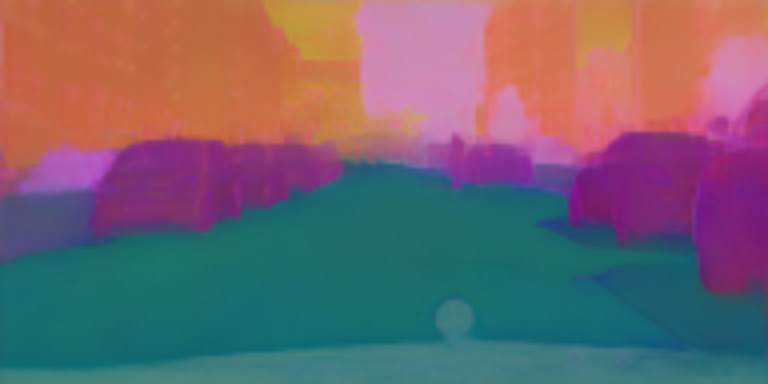}}\\

\bmvaHangBox{\includegraphics[width=1.4cm]{fig/seg/frankfurt_000000_000576_gtFine_color.png}}&
\bmvaHangBox{\includegraphics[width=1.4cm]{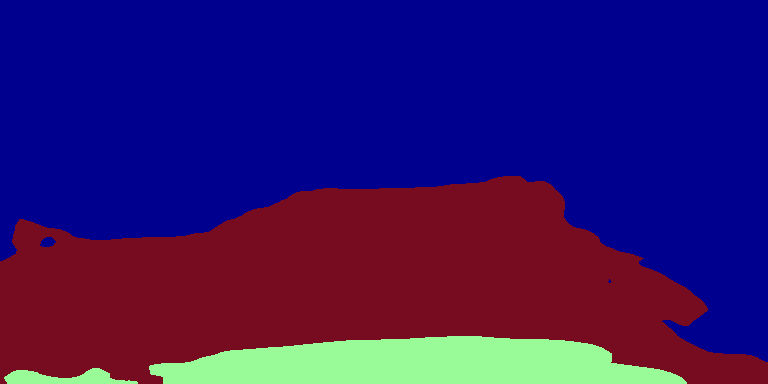}}&
\bmvaHangBox{\includegraphics[width=1.4cm]{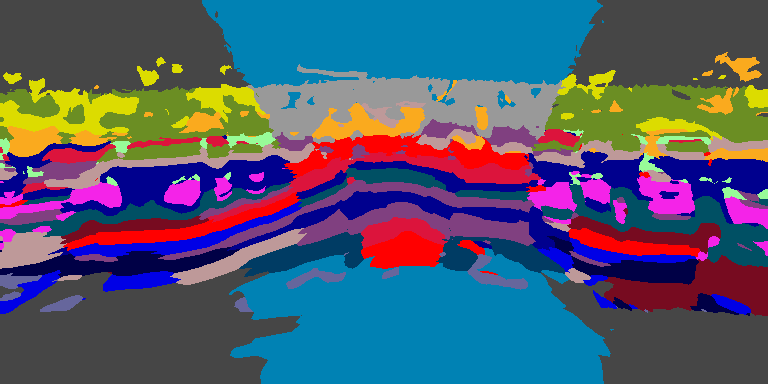}}&
\bmvaHangBox{\includegraphics[width=1.4cm]{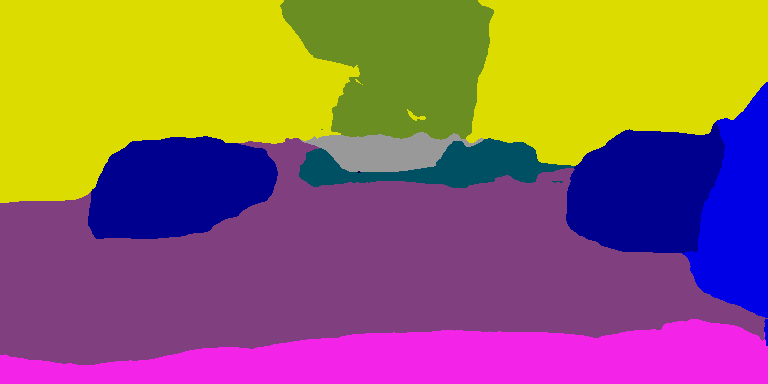}}&
\bmvaHangBox{\includegraphics[width=1.4cm]{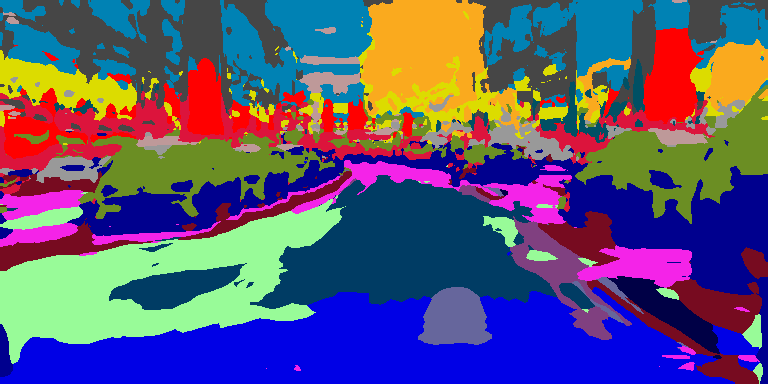}}&
\bmvaHangBox{\includegraphics[width=1.4cm]{fig/seg/depth_mix.png}}&
\bmvaHangBox{\includegraphics[width=1.4cm]{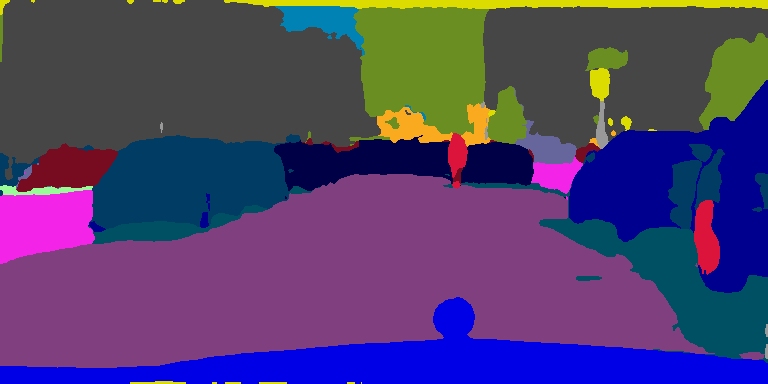}}\\

\bmvaHangBox{\includegraphics[width=1.4cm]{fig/pca/frankfurt_000000_001016_leftImg8bit.png}}&
\bmvaHangBox{\includegraphics[width=1.4cm]{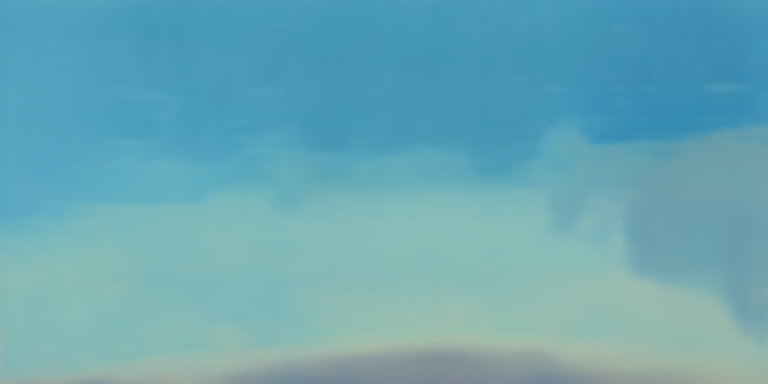}}&
\bmvaHangBox{\includegraphics[width=1.4cm]{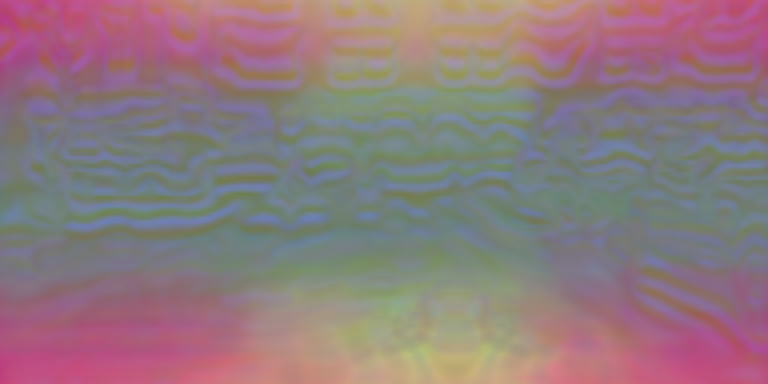}}&
\bmvaHangBox{\includegraphics[width=1.4cm]{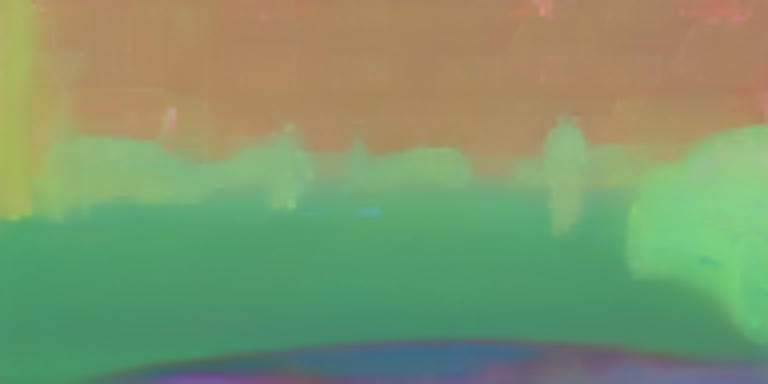}}&
\bmvaHangBox{\includegraphics[width=1.4cm]{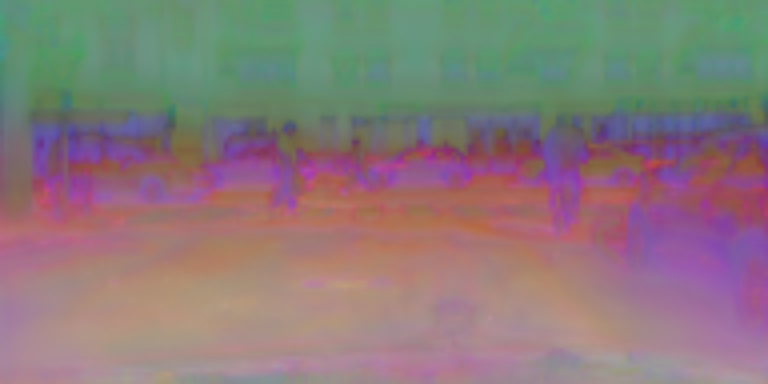}}&
\bmvaHangBox{\includegraphics[width=1.4cm]{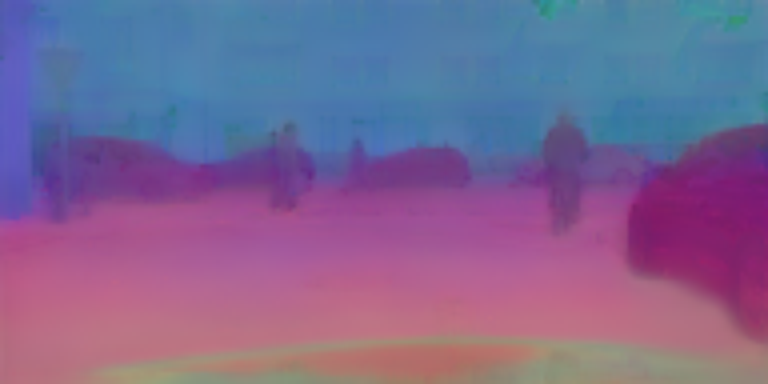}}&
\bmvaHangBox{\includegraphics[width=1.4cm]{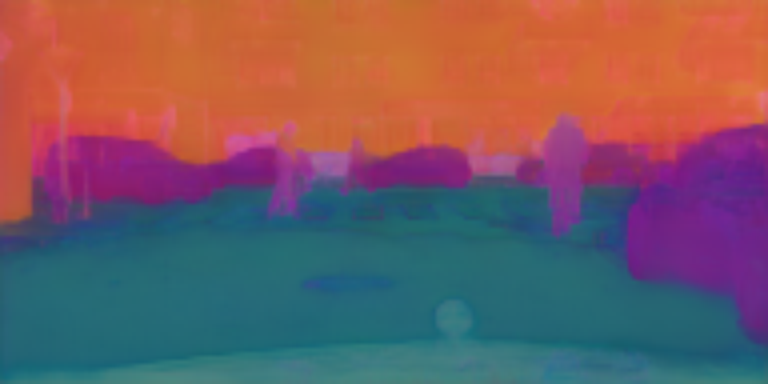}}\\

\bmvaHangBox{\includegraphics[width=1.4cm]{fig/seg/frankfurt_000000_001016_gtFine_color.png}}&
\bmvaHangBox{\includegraphics[width=1.4cm]{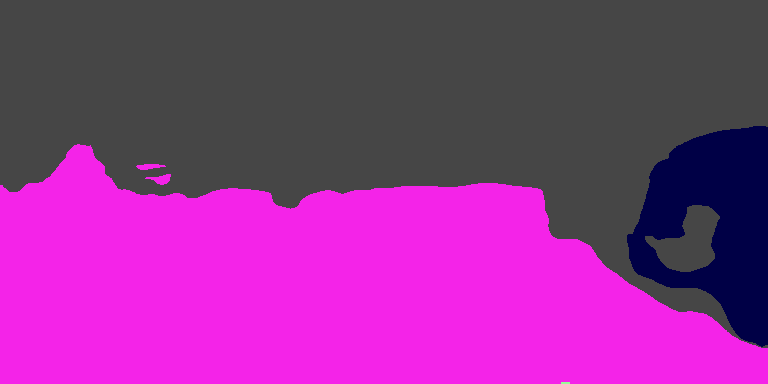}}&
\bmvaHangBox{\includegraphics[width=1.4cm]{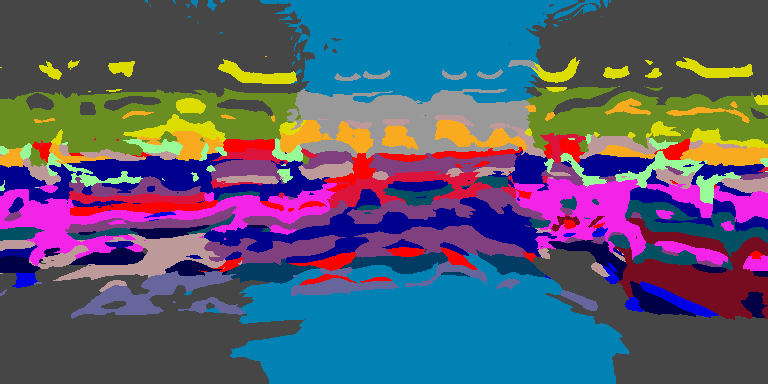}}&
\bmvaHangBox{\includegraphics[width=1.4cm]{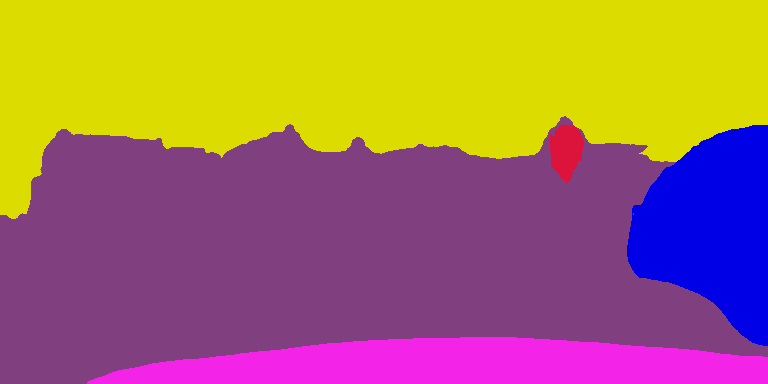}}&
\bmvaHangBox{\includegraphics[width=1.4cm]{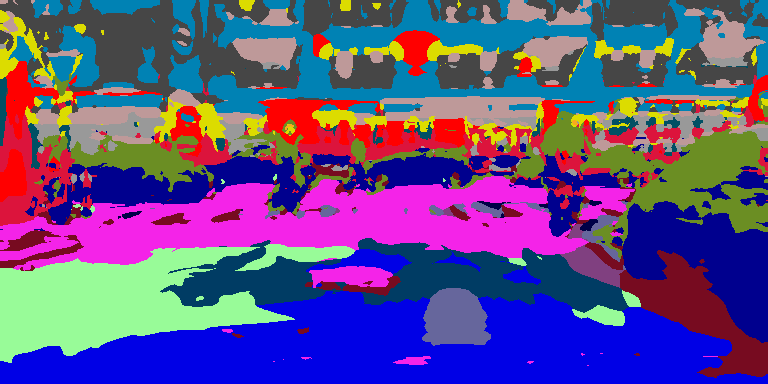}}&
\bmvaHangBox{\includegraphics[width=1.4cm]{fig/seg/depth_mix_2.png}}&
\bmvaHangBox{\includegraphics[width=1.4cm]{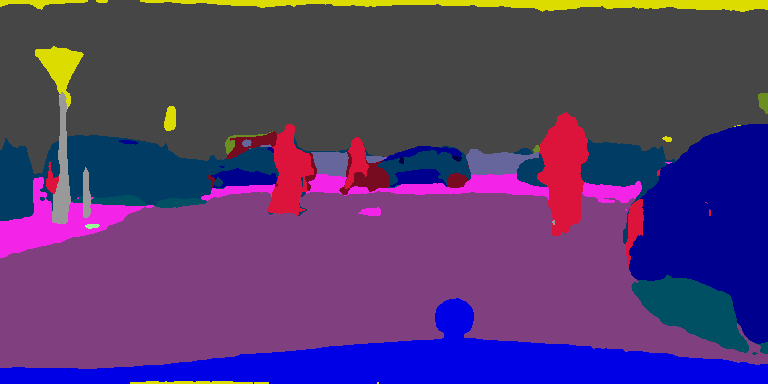}}\\

Images& $\lambda=0$ & $\lambda=1$ & $\lambda=0$ & $\lambda=1$ & $\lambda=0.5$ & $\lambda=0.5$\\
& w/o CP & w/o CP & CP & CP & CP & CP(GT)\\

\end{tabular}
\caption{We visualize the feature maps as RGB images by PCA reduction~\cite{DBLP:conf/eccv/VondrickSFGM18} and the unsupervised semantic segmentation result by representation clustering. $\lambda$ is the loss weight that balance the loss from the two types of positive samples in Eqn.~\ref{L}. We use our coherent depth region proposal for copy-paste except the last column using the ground truth proposal as the upper bound. }
\label{v1}
\end{figure}

\begin{figure}[H]
\centering
\begin{tabular}{lccc}
\centering
Region Proposal &$\lambda=0$&$\lambda=0.5$&$\lambda=1$\\
Ours&\bmvaHangBox{\includegraphics[width=2.5cm]{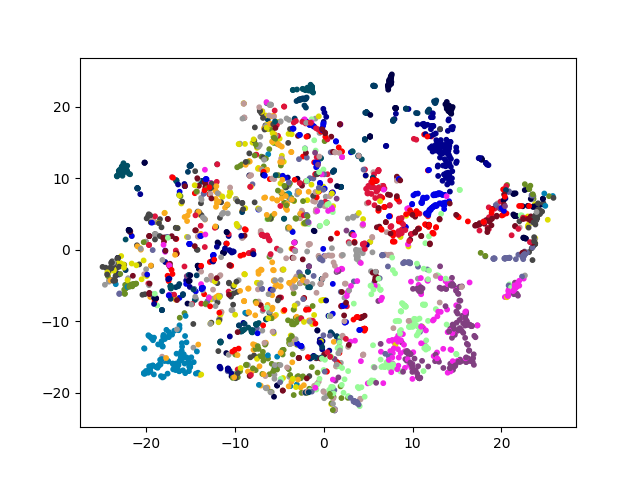}}&
\bmvaHangBox{\includegraphics[width=2.5cm]{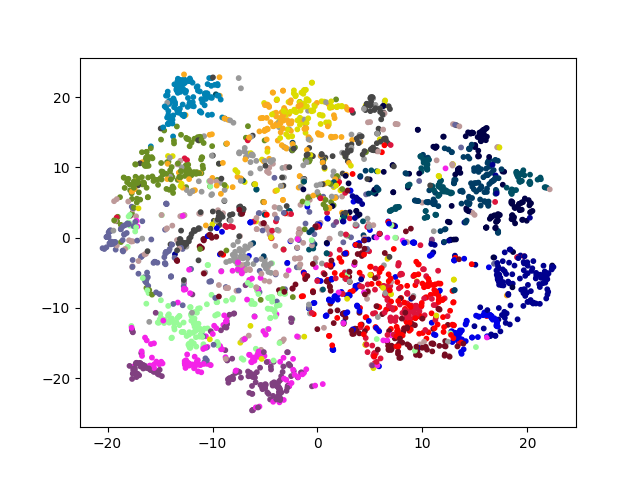}}&
\bmvaHangBox{\includegraphics[width=2.5cm]{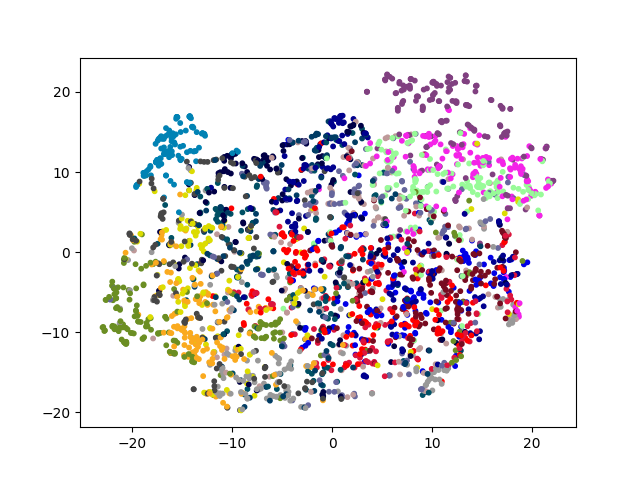}}\\
GT&\bmvaHangBox{\includegraphics[width=2.5cm]{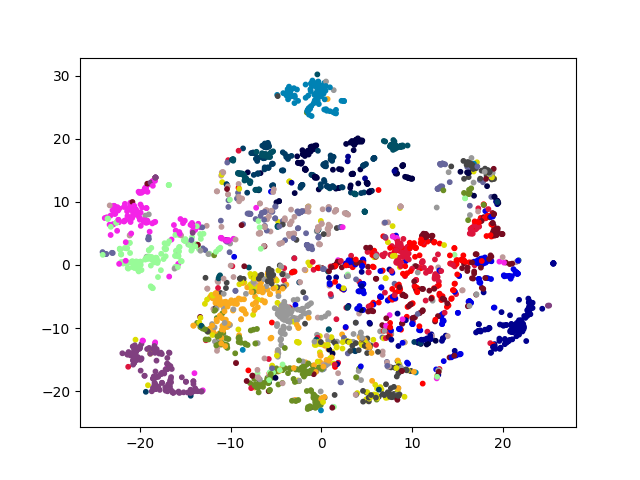}}&
\bmvaHangBox{\includegraphics[width=2.5cm]{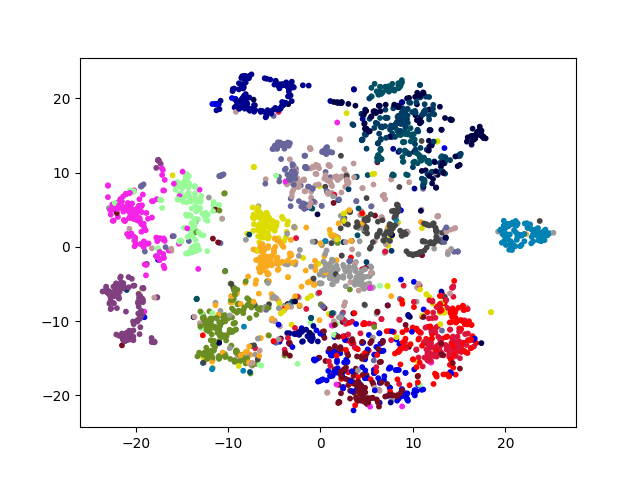}}&
\bmvaHangBox{\includegraphics[width=2.5cm]{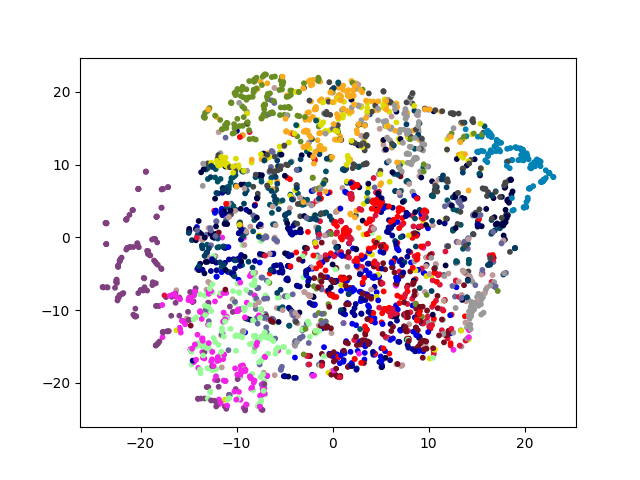}}\\

\end{tabular}
\caption{We visualize the distribution of 3000 sample representations by t-sne. Samples are equally sampled from different classes and colored according to their true classes. Copy-paste is enabled using our coherent depth regions or ground truth mask. $\lambda$ is the loss weight that balance the loss from the two types of positive samples in Eqn.~\ref{L}. We can see that the combination of pixel-level and region-level positive samples helps the clustering of representations from the same class. }
\label{v2}
\end{figure}

We provide more visualizations based on representation map along with input images in Figure \ref{v1} and representation distribution w.r.t. true class labels in Figure \ref{v2}. Copy-pasting coherent depth regions and the combination of pixel-level and region-level positive samples encourage the learning of rich, context-invariant and object-specific representations for unsupervised clustering. We can also see that there is a gap between our coherent depth regions and GT. Especially, region-level positive samples are more sensitive to proposal quality, which may explain why enabling region-level positive samples takes little effect on fine-tuning performance. Future work may explore how to better extract coherent depth regions such as using data-driven methods.

\end{document}